\pdfoutput=1

\documentclass[11pt]{article}

\usepackage[]{EACL2023}

\usepackage{amsmath}
\usepackage{amssymb}
\usepackage{multirow}
\usepackage{multicol}
\usepackage{xcolor}
\usepackage{color}
\usepackage{times}
\usepackage{latexsym}
\usepackage{graphicx}
\usepackage{algorithm}
\usepackage{algpseudocode}
\usepackage{listings}
\usepackage{amssymb}
\usepackage{pifont}
\newcommand{\cmark}{\ding{51}}%
\newcommand{\xmark}{\ding{55}}%

\usepackage[T1]{fontenc}

\usepackage[utf8]{inputenc}

\usepackage{microtype}

\usepackage{inconsolata}

\usepackage{booktabs}
\usepackage{ragged2e}
\usepackage{mathtools}
\usepackage{xspace} 

\newcommand{\ours}{PR-MCS}
\newcommand{\dataset}{M-FineCapEval\xspace}

\title{PR-MCS: Perturbation Robust Metric for MultiLingual Image Captioning}

\author{Yongil Kim$^{1}$, Yerin Hwang$^{1}$, Hyeongu Yun$^{1}$,\\\bf Seunghyun Yoon$^{2}$, 
 Trung Bui$^{2}$, ~\and Kyomin Jung$^{1}$ \\
  $^{1}$Dept. of Electrical and Computer Engineering, Seoul National University \\
  $^{2}$Adobe Research\\
  \texttt{\{miles94, dpfls589, youaredead\}@snu.ac.kr},
  \texttt{\{syoon, bui\}@adobe.com},
  \texttt{kjung@snu.ac.kr}
  }

\begin{document}
\maketitle

\begin{abstract}
Vulnerability to lexical perturbation is a critical weakness of automatic evaluation metrics for image captioning.
This paper proposes \textbf{P}erturbation \textbf{R}obust \textbf{M}ulti-Lingual \textbf{C}LIP\textbf{S}core(\textbf{PR-MCS}), which exhibits robustness to such perturbations, as a novel reference-free image captioning metric applicable to multiple languages.
To achieve perturbation robustness, we fine-tune the text encoder of CLIP with our language-agnostic method to distinguish the perturbed text from the original text.
To verify the robustness of PR-MCS, we introduce a new fine-grained evaluation dataset consisting of detailed captions, critical objects, and the relationships between the objects for $3,000$ images in five languages\footnote{All the datasets and code will be available here. Our dataset is distributed under the CC-BY-NC 4.0 license.}.
In our experiments, PR-MCS significantly outperforms baseline metrics in capturing lexical noise of all various perturbation types in all five languages, proving that PR-MCS is highly robust to lexical perturbations.
\end{abstract}

\section{Introduction}
\label{sec:introduction}
Image captioning~\cite{xu2015show, vinyals2015show, vinyals2016show, lu2017knowing} is a multimodal task that automatically generates captions that describe the visual content of an image and integrates multiple disciplines of visual and textual modality.
Image captioning is a natural language generation (NLG) task~\cite{gatt2018survey}, but the evaluation metric has different characteristics from other NLG metrics~\cite{sai2022survey}.
Image captioning metrics should be able to evaluate not only linguistic fluency and syntactic thoroughness but also semantic correspondence to visual content\cite{bai2018survey}.

\begin{figure}[h]
\centering
\includegraphics[width=1.0\columnwidth]{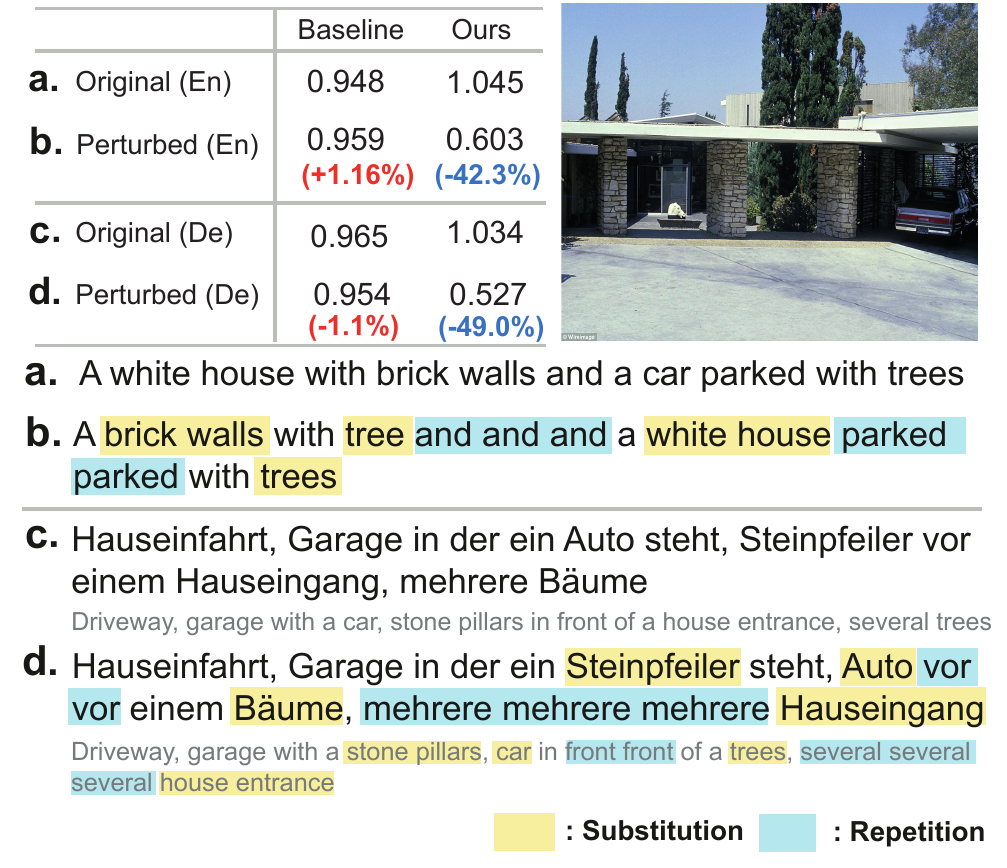} 
\caption{An example for perturbation robustness test. The baseline metric shows similar scores to both original and perturbed captions in English and German, but our metric shows significant score drop for perturbed captions indicating that perturbation is well detected. Translated captions of German to English are provided for explanation purposes.}
\label{fig1}
\end{figure}

Evaluation criteria for image captioning have evolved from N-gram-based metrics \cite{papineni2002bleu, lin2004rouge, banerjee2005meteor, vedantam2015cider} to reference-free metrics~\cite{lee2020vilbertscore, lee2021umic, hessel2021clipscore}.
Recently, CLIPScore \cite{hessel2021clipscore} has been proposed to leverage the large-scale pretrained vision and language model CLIP~\cite{radford2021learning}.
In evaluating generated captions by computing cosine similarity between embedded vectors (i.e., image and text) using CLIP, CLIPScore achieves a higher correlation with human judgments than traditional metrics.

However, \citet{sai2021perturbation} have revealed that current metrics are prone to failure in capturing lexical noise in generated captions.
For example, when a perturbation is applied to an original caption (e.g., a removal or swap at the token level), existing image captioning metrics do not recognize the change and compute a score similar to that for the original caption case.
This failure to capture lexical noise raises a critical question concerning the reliability of the metric, as shown in the example in Figure~\ref{fig1}.
CLIPScore exhibits the same tendency in our analysis, which reflects its vulnerability to perturbed texts.
By extending CLIPScore to a multilingual setting, we observe that a multilingual CLIPScore exhibits the same limitations in multiple languages other than English, i.e., French, German, Spanish, and Japanese.

In this paper, we address this problem by proposing a novel method for enhancing the perturbation robustness of CLIPScore.
Our method is to fine-tune the text encoder of CLIP with perturbed captions so that the text encoder can distinguish the perturbed text embeddings from the original text embeddings.
The simplicity and effectiveness of our method enable us to apply it to multiple languages without relying on human annotations.
Using our method, we develop Perturbation-Robust Multilingual CLIPScore (PR-MCS), a perturbation-robust and language-agnostic metric for image captioning.
Experimental results show that PR-MCS outperforms baseline metrics in capturing lexical noise in captions in all five languages considered.

Furthermore, we create a new multilingual fine-grained caption evaluation dataset, M-FineCapEval, to validate the perturbation robustness of PR-MCS.
Currently, most of the image captioning evaluation datasets are limited to  English, so a translation process using machine translation~\cite{bahdanau2014neural, johnson2017google} is essential for multilinguality experiments.
However, this process depends on the performance of the translation model, so there is a disadvantage in that the evaluation reliability is low compared to human-annotated labels.
Accordingly, we ask human experts to describe captions in the five languages mentioned above. In addition, to provide various types of effective lexical noise to the caption, the caption was configured as fine-grained, and the experts were instructed to point out the critical objects in the caption.
We have made our code publicly available\footnote{https://github.com/yong1-kim/PR-MCS}.

Our contributions are as follows:
\begin{itemize}
\item We develop a language-agnostic method for fine-tuning the CLIP text encoder to capture lexical perturbations.
\item We propose PR-MCS, a perturbation-robust metric for image captioning in multiple languages.
\item We introduce a fine-grained multilingual image captioning evaluation set, M-FineCapEval, to demonstrate the performance and robustness of our method.
\end{itemize}

\section{Related works}
\label{sec:related works}
\textbf{Image captioning metrics}
As with other natural language generation tasks, image captioning can be evaluated using various proposed metrics.
BLEU~\cite{papineni2002bleu}, ROUGE~\cite{lin2004rouge}, and METEOR~\cite{banerjee2005meteor} are representative image captioning metrics based on n-gram similarity with reference captions.
Other widely used reference-based metrics include CIDEr~\cite{vedantam2015cider}, which weights n-gram similarity~\cite{kondrak2005n} through TF-IDF~\cite{aizawa2003information}, and SPICE~\cite{anderson2016spice}, which evaluates captioning based on scene graphs.
Recently, reference-based metrics using embedding similarity with reference captions based on a model, such as BERTScore~\cite{zhang2019bertscore}, BERT-TBR~\cite{yi2020improving}, and VilBERTScore~\cite{lee2020vilbertscore}, have been introduced.

\indent Researchers have also proposed unreferenced image captioning metrics that evaluate generated captions by comparing them with original images.
For instance, VIFIDEL~\cite{madhyastha2019vifidel} uses the word mover distance~\cite{kusner2015word} between the image and candidate caption, and UMIC~\cite{lee2021umic}, which fine-tunes UNITER~\cite{chen2020uniter} using contrastive loss from augmented captions, directly evaluates captions generated from vision-and-language embedding spaces.
\\
\\
\noindent \textbf{CLIPScore} 
CLIPScore~\cite{hessel2021clipscore} is a reference-free metric that does not require ground-truth captions.
CLIPScore relies heavily on the CLIP~\cite{radford2021learning} model, trained with 400 million image caption pairs using a contrastive objective function that distinguishes original image–caption pairs from unmatched captions.
The calculated CLIPScore is the weighted value of cosine similarity between image embedding and text embedding encoded by the CLIP model.
Thanks to the power of the massive training dataset and CLIP’s objective function, CLIPScore exhibits a high correlation with human evaluation.
However, CLIPScore is limited in that it is an image captioning metric that applies only to English.
In this study, we propose a new multilingual image captioning metric developed by extending CLIPScore to a multilingual setting.\\

\noindent \textbf{Perturbation Robustness}
In a recent study,  \citet{sai2021perturbation} selected various criteria for use in assessing how various NLG evaluation metrics perform.
In addition, perturbation was applied to multiple image captioning factors to assess the perturbation robustness of the image captioning metrics.
 \citet{sai2021perturbation} provided a perturbation checklist of metrics for NLG tasks; we go further and present a novel metric that overcomes the limitations of other metrics.
We select some perturbation criteria from among those suggested by  \citet{sai2021perturbation}, designate them as target perturbations, and show that the CLIPScore cannot detect these perturbations in multiple languages.
Even if the generated captions are corrupted, CLIPScore outputs similar results for the original and corrupted sentences.
This study proposes a novel metric with perturbation robustness based on CLIPScore to address its weaknesses in multiple languages.

\begin{figure}[t]
\centering
\includegraphics[width=1.0\columnwidth]{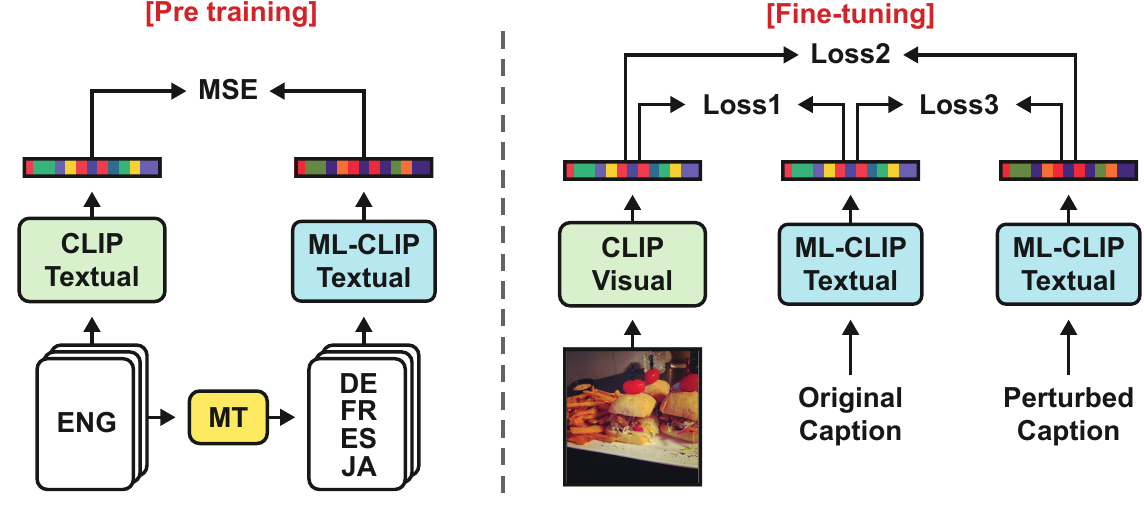} 
\caption{Overall training procedure of PR-MCS. We pre-train Multilingual text encoder with teacher learning. Then, we fine-tune the Multilingual text encoder.}
\label{fig2}
\end{figure}
\section{Perturbation Robust Multi-Lingual CLIPScore}
\label{sec:methods}

\subsection{MultiLingual CLIPscore}
\label{MCS}
We propose a new multilingual image captioning metric, \textbf{M}ultilingual \textbf{C}LIP\textbf{S}core (MCS), to overcome the limitation of CLIPScore that it can only be applied to English.
Recently, \citet{multilingualCLIP2022} proposed a multilingual CLIP model applicable to various languages by learning the expressive power of CLIP’s text encoder through teacher learning.
Similarly, in this study, we pre-train the multilingual text encoder with teacher learning (Figure~\ref{fig2} left).
In teacher learning, the multilingual text encoder learns the pre-trained CLIP text embedding of an English sentence so that the embeddings of the sentence translated through the machine translation model are similar.
We use MSE loss as the teacher learning loss, and the formula is as follows:
$$ L(t,s) = \frac{1}{N}\sum_{i=0}^{n}(t-s)^{2}, $$

\noindent where $t$ is the teacher embedding and $s$ is the student embedding. More details on multilingual textual encoder pre-training can be found in the appendix~\ref{pretraining_details}.

We present a new multilingual image captioning metric, MCS, using this model as a backbone.
MCS uses image–caption pairs with weight given to the cosine similarity of embeddings created through visual and text encoders, respectively.
The formula for an image–caption input pair $(I, c)$ in MCS is as follows:
$$MCS {(I, c) = w * max( 0, cos(V(I), T(c))},$$

\noindent where $V(I)$ is the visual embedding where the image is passed through the visual encoder and $T(c)$ is the text embedding where the caption is passed through the multilingual text encoder. 
The value of $w$ is set to 2.5, as in the original CLIPScore paper.

\begin{figure}[t]
\centering
\includegraphics[width=0.8\columnwidth]{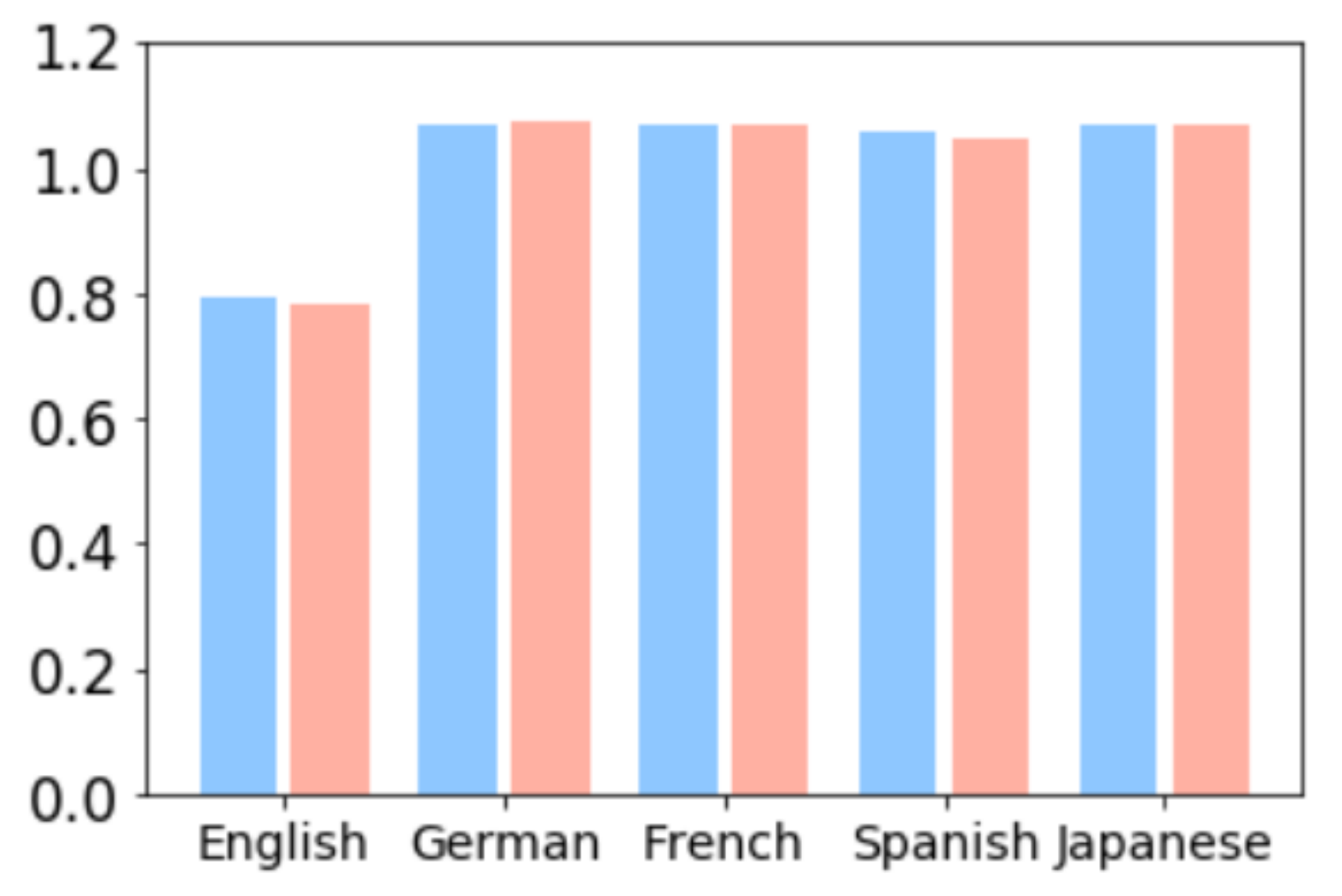} 
\caption{Scores of original captions (blue) and perturbed captions (red) with repetitive lexical noises. In all cases, the perturbed captions show no differences from the original captions.}
\label{fig3}
\end{figure}

\subsection{Vulnerability to lexical perturbation}
We employed some of the perturbation criteria identified by \citet{sai2021perturbation} and checked the perturbation sensitivity of CLIPScore and MCS.
One of the criteria, \textit{"Repetition"}, is a perturbation in which words appear repeatedly at the token level in the original caption (e.g., “\textit{I am a boy.}” $\rightarrow$ “\textit{I am am a boy boy.}”).
The figure~\ref{fig3} shows what score is given by baseline metrics when repetitive lexical noise is introduced.
We randomly selected 3,000 samples from the MSCOCO~\cite{lin2014microsoft} dataset, translated them into four languages, and gave repetitive lexical noise to the captions.
The blue bars are the score for the original captions, and the red bars are the score for the perturbed captions.
For English, CLIPScore is used as a metric, and for other languages, MCS is used for score extraction.
The caption to which lexical noise is added is expected to have a lower matching score with the image than the original caption.
However, for all languages, the scores for the perturbed captions are not lower than those for the original captions. There are even cases in which the perturbed caption is given a higher score.
Similar tendencies can be observed for other perturbation criteria as well as \textit{"Repetition"}.
These results confirm that CLIPScore and MCS are limited in that they are vulnerable to lexical perturbation and that a metric that is robust to perturbation is needed.

\subsection{Perturbation-Robust Multilingual CLIPScore (PR-MCS)}
We introduce a novel language-agnostic perturbation method that increases the robustness of MCS.
This method of fine-tuning the multilingual text encoder is to add three losses to original CLIP loss.
The CLIPScore is constructed through embeddings of pre-trained CLIP without additional training. 
The CLIP loss $\mathcal{L}_{CLIP}$ consists of in-batch contrastive loss using cross-entropy loss, and the implementation is the same as the pre-training loss of CLIP.
We construct a loss based on the contrastive loss of CLIP to maintain the high correlation with the human judgment of CLIPScore.

Then, we train the text encoder by adding three additional losses for perturbation robustness. 
These losses aim to maintain the close relationship between the image embedding and the original caption while increasing the distance from the perturbed caption.
An (image, original caption, perturbed caption) triplet is then used as input to fine-tune the text encoder through three losses, as shown in Figure~\ref{fig2} (right). The losses are as follows:
\begin{align} 
    \mathcal{L}_{1}& = 1 - cos(V(I), T(o)) ,\label{equation:finetuning_loss1}\\
    \mathcal{L}_{2}& = max(0, cos(V(I), T(p)) ,\label{equation:finetuning_loss2}\\
    \mathcal{L}_{3}& = max(0, cos(T(o), T(p)), \label{equation:finetuning_loss3}
\end{align}
\noindent where $(I,o,p)$ is the (image, original caption, perturbed caption) triplet, $V$ is a visual encoder, and $T$ is a text encoder. 

Equation (\ref{equation:finetuning_loss1}) is composed of the cosine embedding loss of the two representations needed to increase the similarity between the image embedding and the original caption.
Since MCS is based on cosine similarity, the purpose of Eq. (\ref{equation:finetuning_loss1}) is to obtain a higher score for the original caption.
Equation (\ref{equation:finetuning_loss2}) reduces the cosine similarity of image embedding and perturbed caption embedding.
Equation (\ref{equation:finetuning_loss3}) reduces the similarity between the original and perturbed caption embeddings.
These three losses are combined to obtain the final objective function as follows:
$$\mathcal{L} = \mathcal{L}_{CLIP} + \lambda_{1} * \mathcal{L}_{1} +  \lambda_{2} * \mathcal{L}_{2} + \lambda_{3} * \mathcal{L}_{3}.$$

\noindent We develop PR-MCS by fine-tuning Multilingual CLIP using the proposed loss function.
$$PR {\text -} MCS {(I, c) = w * max( 0, cos(V(I), T^{*}(c))},$$
\noindent where $T^{*}(c)$ is the text embedding from the fine-tuned multi-lingual text encoder. 
$w$ is also set to 2.5, as in the original CLIPScore and MCS.

\begin{figure}[t]
\centering
\includegraphics[width=1.0\columnwidth]{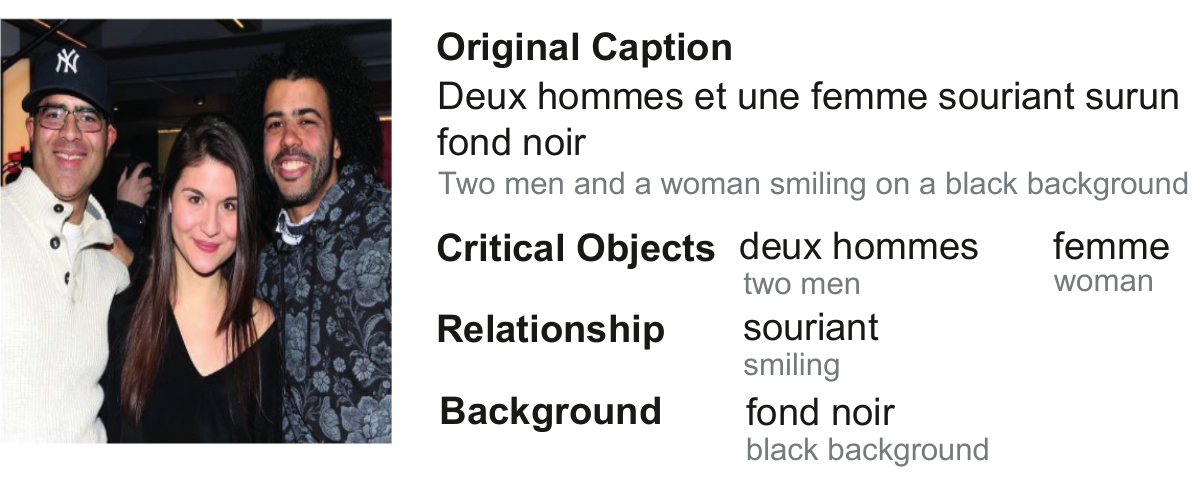} 
\caption{An example of the proposed \dataset~dataset (translation provided for explanation purpose). \dataset~consists of 3,000 samples of fine-grained captions, critical objects, backgrounds, and relationships.}
\label{fig4}
\end{figure}

\section{\dataset~ Dataset}
\label{sec:Mfinecapeval}
To evaluate the performance of PR-MCS, we introduce a new evaluation set, \dataset. Most existing image captioning datasets are limited to English~\cite{young2014image, krishna2017visual}. Therefore, a model for machine translation (MT) from English to other languages is essential to evaluate image captioning in various languages. However, translated evaluation set is highly dependent on the performance of the MT model and is highly likely to have an English-language bias. In addition, translation is inarticulate in the target language because it is difficult to reflect the unique characteristics of each language, such as word order and lexical choice~\cite{zhang2019effect, cao2020multilingual}. Therefore, the results obtained using a translated evaluation set achieve poorer agreement with human evaluation than those obtained using the English evaluation set. For these reasons, a human-annotated image captioning evaluation set with a wide variety of languages is needed.\\

\noindent \textbf{Multilingual image captioning evaluation set} 
We introduce a new human-annotated multilingual evaluation dataset, \dataset. We extended FineCapEval~\cite{cho2022fine}, which has only English captions, to five languages (English, German, French, Spanish, and Japanese).
Human experts viewed the images for each language and added captions directly.
Each sentence generated directly by native speakers is more fluent than translated versions.
Moreover, \dataset can capture various cultural aspects that MT models cannot~\cite{liu2021visually}.
Therefore, the reliability of evaluation in multilingual settings increases. An example of the dataset is shown in Figure~\ref{fig4}. Human annotators for each language created a caption, critical objects, backgrounds, and relationships for a given image.\\

\begin{table}[t]

\renewcommand{\arraystretch}{1.5} 
\resizebox{\columnwidth}{!}{
\begin{tabular}{c|ccccc}
\hline
\textbf{Language}                                                                     & \textbf{En} & \textbf{De} & \textbf{Es} & \textbf{Fr} & \textbf{Ja} \\ \hline
\textbf{Datasize}                                                                     & 1,000        & 3,000        & 3000        & 3,000        & 3,000        \\ \hline
\textbf{\begin{tabular}[c]{@{}c@{}}Sentence length\end{tabular}}         & 23.42      & 19.42      & 23.21      & 22.09      & 48.39       \\ \hline
\textbf{\begin{tabular}[c]{@{}c@{}}\# of critical objects \end{tabular}} & 2.87       & 3.56       & 4.17       & 3.11       & 4.42       \\ \hline
\textbf{\begin{tabular}[c]{@{}c@{}}\# of backgrounds\end{tabular}}      & 1.25        & 1.27       & 1.39       & 1.25       & 1.56       \\ \hline
\textbf{\begin{tabular}[c]{@{}c@{}}\# of relationships\end{tabular}}    & 1.57       & 2.05       & 1.78       & 1.81       & 2.46       \\ \hline
\end{tabular}
}
\caption{\dataset~statistics}
\label{table1}
\end{table}

\noindent \textbf{Fine-grained caption with critical objects}
To generate various perturbed captions for evaluation, \dataset is constructed with long fine-grained captions of 20 words or more. In addition, we had human experts point to critical objects, backgrounds, and relationships to create perturbed captions effectively. As described above, the image captioning metric should also reflect the semantic correspondence of whether the caption captures information contained in visual content well. The critical object of the caption should point to the most important object of this visual content, so it plays a key role in the comparison between embeddings. When perturbation is applied to this critical object, a more powerful and effective perturbation is achieved.

However, the well-known weakness of CLIP is that it does not produce different results when the positions of critical objects in the sentence are changed. For example, the CLIP text embeddings of the two sentences “\textit{A \textbf{blue} car in front of the \textbf{white} church}” and “\textit{A \textbf{white} car in front of the \textbf{blue} church}” are almost identical. To evaluate the robustness to this perturbation, we construct perturbation criteria using critical object information.\\

\noindent \textbf{Statistics}
Table~\ref{table1} provides detailed statistics for \dataset, including the dataset size for each language, the average sentence length, and the average numbers of critical objects, backgrounds, and relationships. M-FineCapEval consists of lengthy sentences of approximately 20 word tokens on average. In the case of Japanese, since there is no spacing in a sentence, sentence length is calculated using a tokenizer based on word extractor, and the sentence length is almost the same as the character level. In addition, there are three to four critical objects in all languages, so each sentence describes the visual content of an image in great detail.

\begin{figure}[t]
\renewcommand{\arraystretch}{1.5} 
\centering
\resizebox{\columnwidth}{!}{%
\begin{tabular}{|c|l|}
\hline
\textbf{Critical Objects} & \textcolor{red}{\textbf{white shirt, grey shorts, golf, green field}}                                                                                                                                                                                   \\ \hline
\textbf{Original}         & \begin{tabular}[c]{@{}l@{}}A man, wearing a \textcolor{red}{\textbf{white shirt}} and \textcolor{red}{\textbf{grey shorts}}, is playing \\ \textcolor{red}{\textbf{golf}} on a \textcolor{red}{\textbf{green field}} with green trees and a blue sky in\\ the background\end{tabular}                                                             \\ \hline
\textbf{Jumble}           & \begin{tabular}[c]{@{}l@{}}with green playing blue trees a background. green in\\ shorts, and white the is wearing man, a A and grey on sky a\\ golf shirt field\end{tabular}                                                             \\ \hline
\textbf{Removal}          & man, white and shorts, is playing a green trees a                                                                                                                                                                                         \\ \hline
\textbf{Repetition}       & \begin{tabular}[c]{@{}l@{}}A man, man, wearing a white white shirt and and\\ grey shorts, shorts, is is playing playing golf on a a green field\\ with green green trees and a a blue sky in the background.\end{tabular}                 \\ \hline
\textbf{Masking}          & \begin{tabular}[c]{@{}l@{}}A {[}MASK{]} wearing a {[}MASK{]} shirt {[}MASK{]} grey\\ {[}MASK{]} {[}MASK{]} {[}MASK{]} golf on {[}MASK{]} green field with\\ {[}MASK{]} {[}MASK{]} and {[}MASK{]} blue sky in the background.\end{tabular} \\ \hline
\textbf{Substitution}      & \begin{tabular}[c]{@{}l@{}}A man, wearing a \textcolor{red}{\textbf{golf}} and \textcolor{red}{\textbf{green field}}, is playing\\ \textcolor{red}{\textbf{white shirt}} on a \textcolor{red}{\textbf{grey shorts}} with green trees and a blue sky in\\ the background.\end{tabular}                                                             \\ \hline
\end{tabular}%
}
\caption{Example of perturbed captions of \dataset in English. The critical objects are shuffled for in-sentence substitution.}
\label{fig5}
\end{figure}

\begin{figure*}[t]
\centering
\includegraphics[width=1.0\textwidth]{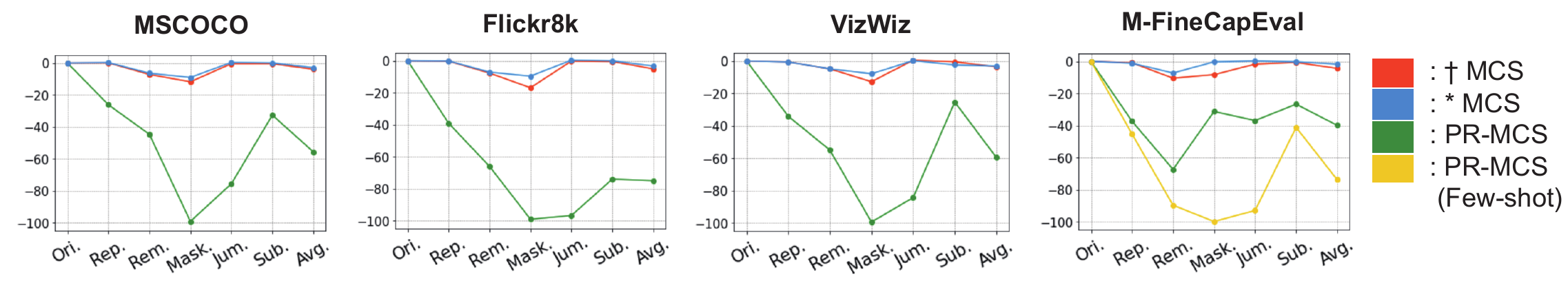} 
\caption{Experiment result graph. The y-axis value represents the score drop of the perturbed caption as a percentage difference compared to the original caption. The experiment is conducted with four datasets, and we report the average of the languages to confirm the results for each perturbation. As a result, it can be seen that PR-MCS is more robust than baseline metrics for all perturbations across all datasets.}
\label{fig6}
\end{figure*}

\section{Experiments}
\label{sec:experiments}
Our framework seeks to identify whether a given metric can detect lexical noise in a generated caption. Through exhaustive experiments, we evaluate whether the PR-MCS developed as described in this paper successfully distinguishes the perturbed caption from the original caption.

\begin{table*}[t]
\renewcommand{\arraystretch}{1.3}
\centering
\resizebox{0.9\textwidth}{!}{%
\begin{tabular}{cccccccccccc}
\hline
\multicolumn{1}{c|}{\multirow{2}{*}{\textbf{Eval Dataset}}}  & \multicolumn{1}{c|}{\multirow{2}{*}{\textbf{Metric}}}                                                      & \multicolumn{2}{c|}{\textbf{EN}}                                                                                    & \multicolumn{2}{c|}{\textbf{DE}}                                                                                    & \multicolumn{2}{c|}{\textbf{FR}}                                                                                    & \multicolumn{2}{c|}{\textbf{ES}}                                                                                    & \multicolumn{2}{c}{\textbf{JA}}                                                                \\ \cline{3-12} 
\multicolumn{1}{c|}{}                                        & \multicolumn{1}{c|}{}                                                                                      & \textbf{Original}       & \multicolumn{1}{c|}{\textbf{\begin{tabular}[c]{@{}c@{}}Perturbed\\ Average\end{tabular}}} & \textbf{Original}       & \multicolumn{1}{c|}{\textbf{\begin{tabular}[c]{@{}c@{}}Perturbed\\ Average\end{tabular}}} & \textbf{Original}       & \multicolumn{1}{c|}{\textbf{\begin{tabular}[c]{@{}c@{}}Perturbed\\ Average\end{tabular}}} & \textbf{Original}       & \multicolumn{1}{c|}{\textbf{\begin{tabular}[c]{@{}c@{}}Perturbed\\ Average\end{tabular}}} & \textbf{Original}       & \textbf{\begin{tabular}[c]{@{}c@{}}Perturbed\\ Average\end{tabular}} \\ \hline
\multicolumn{1}{c|}{\textbf{}}                               & \multicolumn{1}{c|}{\multirow{2}{*}{\textbf{$\dagger$MCS}}}                                                         & \multirow{2}{*}{0.7944} & \multicolumn{1}{c|}{0.7442}                                                               & \multirow{2}{*}{1.0743} & \multicolumn{1}{c|}{1.03562}                                                              & \multirow{2}{*}{1.0719} & \multicolumn{1}{c|}{1.03298}                                                              & \multirow{2}{*}{1.0564} & \multicolumn{1}{c|}{1.01878}                                                              & \multirow{2}{*}{1.0727} & 1.03122                                                              \\
\multicolumn{1}{c|}{\textbf{}}                               & \multicolumn{1}{c|}{}                                                                                      &                         & \multicolumn{1}{c|}{(-6.32\%)}                                                            &                         & \multicolumn{1}{c|}{(-3.60\%)}                                                            &                         & \multicolumn{1}{c|}{(-3.63\%)}                                                            &                         & \multicolumn{1}{c|}{(-3.56\%)}                                                            &                         & (-3.87\%)                                                            \\
\multicolumn{1}{c|}{\multirow{2}{*}{\textbf{MSCOCO}}}        & \multicolumn{1}{c|}{\multirow{2}{*}{\textbf{*MCS}}}                                                         & \multirow{2}{*}{1.0939} & \multicolumn{1}{c|}{1.05808}                                                              & \multirow{2}{*}{1.0839} & \multicolumn{1}{c|}{1.05512}                                                              & \multirow{2}{*}{1.0837} & \multicolumn{1}{c|}{1.053}                                                                & \multirow{2}{*}{1.0799} & \multicolumn{1}{c|}{1.04988}                                                              & \multirow{2}{*}{1.0286} & 0.9813                                                               \\
\multicolumn{1}{c|}{}                                        & \multicolumn{1}{c|}{}                                                                                      &                         & \multicolumn{1}{c|}{(-3.27\%)}                                                            &                         & \multicolumn{1}{c|}{(-2.66\%)}                                                            &                         & \multicolumn{1}{c|}{(-2.83\%)}                                                            &                         & \multicolumn{1}{c|}{(-2.78\%)}                                                            &                         & (-4.60\%)                                                            \\
\multicolumn{1}{c|}{\textbf{}}                               & \multicolumn{1}{c|}{\multirow{2}{*}{\textbf{PR-MCS}}}                                                      & \multirow{2}{*}{1.4177} & \multicolumn{1}{c|}{\textbf{0.29964}}                                                     & \multirow{2}{*}{1.3153} & \multicolumn{1}{c|}{\textbf{0.51562}}                                                     & \multirow{2}{*}{1.4127} & \multicolumn{1}{c|}{\textbf{0.66434}}                                                     & \multirow{2}{*}{1.4086} & \multicolumn{1}{c|}{\textbf{0.6814}}                                                      & \multirow{2}{*}{1.3948} & \textbf{0.92022}                                                     \\
\multicolumn{1}{c|}{\textbf{}}                               & \multicolumn{1}{c|}{}                                                                                      &                         & \multicolumn{1}{c|}{\textbf{(-78.86\%)}}                                                  &                         & \multicolumn{1}{c|}{\textbf{(-60.80\%)}}                                                  &                         & \multicolumn{1}{c|}{\textbf{(-52.97\%)}}                                                  &                         & \multicolumn{1}{c|}{\textbf{(-51.63\%)}}                                                  &                         & \textbf{(-34.02\%)}                                                  \\ \hline
\multicolumn{1}{c|}{\textbf{}}                               & \multicolumn{1}{c|}{\multirow{2}{*}{\textbf{$\dagger$MCS}}}                                                         & \multirow{2}{*}{1.0659} & \multicolumn{1}{c|}{1.0114}                                                               & \multirow{2}{*}{1.0688} & \multicolumn{1}{c|}{1.01966}                                                              & \multirow{2}{*}{1.0653} & \multicolumn{1}{c|}{1.01066}                                                              & \multirow{2}{*}{1.0527} & \multicolumn{1}{c|}{0.99786}                                                              & \multirow{2}{*}{1.0703} & 1.01272                                                              \\
\multicolumn{1}{c|}{\textbf{}}                               & \multicolumn{1}{c|}{}                                                                                      &                         & \multicolumn{1}{c|}{(-5.11\%)}                                                            &                         & \multicolumn{1}{c|}{(-4.60\%)}                                                            &                         & \multicolumn{1}{c|}{(-5.13\%)}                                                            &                         & \multicolumn{1}{c|}{(-5.21\%)}                                                            &                         & (-5.38\%)                                                            \\
\multicolumn{1}{c|}{\multirow{2}{*}{\textbf{Flickr8k}}}      & \multicolumn{1}{c|}{\multirow{2}{*}{\textbf{*MCS}}}                                                         & \multirow{2}{*}{1.0915} & \multicolumn{1}{c|}{1.0543}                                                               & \multirow{2}{*}{1.0779} & \multicolumn{1}{c|}{1.04554}                                                              & \multirow{2}{*}{1.0758} & \multicolumn{1}{c|}{1.04376}                                                              & \multirow{2}{*}{1.0747} & \multicolumn{1}{c|}{1.04218}                                                              & \multirow{2}{*}{1.0293} & 0.9879                                                               \\
\multicolumn{1}{c|}{}                                        & \multicolumn{1}{c|}{}                                                                                      &                         & \multicolumn{1}{c|}{(-3.41\%)}                                                            &                         & \multicolumn{1}{c|}{(-3.00\%)}                                                            &                         & \multicolumn{1}{c|}{(-2.98\%)}                                                            &                         & \multicolumn{1}{c|}{(-3.03\%)}                                                            &                         & (-4.02\%)                                                            \\
\multicolumn{1}{c|}{\textbf{}}                               & \multicolumn{1}{c|}{\multirow{2}{*}{\textbf{PR-MCS}}}                                                      & \multirow{2}{*}{1.1484} & \multicolumn{1}{c|}{\textbf{0.24016}}                                                     & \multirow{2}{*}{1.1621} & \multicolumn{1}{c|}{\textbf{0.26744}}                                                     & \multirow{2}{*}{1.6681} & \multicolumn{1}{c|}{\textbf{0.42752}}                                                     & \multirow{2}{*}{1.6534} & \multicolumn{1}{c|}{\textbf{0.4685}}                                                      & \multirow{2}{*}{1.6279} & \textbf{0.44632}                                                     \\
\multicolumn{1}{c|}{\textbf{}}                               & \multicolumn{1}{c|}{}                                                                                      &                         & \multicolumn{1}{c|}{\textbf{(-79.09\%)}}                                                  &                         & \multicolumn{1}{c|}{\textbf{(-76.99\%)}}                                                  &                         & \multicolumn{1}{c|}{\textbf{(-74.37\%)}}                                                  &                         & \multicolumn{1}{c|}{\textbf{(-71.66\%)}}                                                  &                         & \textbf{(-72.58\%)}                                                  \\ \hline
\multicolumn{1}{c|}{\textbf{}}                               & \multicolumn{1}{c|}{\multirow{2}{*}{\textbf{$\dagger$MCS}}}                                                         & \multirow{2}{*}{0.7217} & \multicolumn{1}{c|}{0.68536}                                                              & \multirow{2}{*}{1.0284} & \multicolumn{1}{c|}{0.99504}                                                              & \multirow{2}{*}{1.0266} & \multicolumn{1}{c|}{0.9925}                                                               & \multirow{2}{*}{1.0226} & \multicolumn{1}{c|}{0.98982}                                                              & \multirow{2}{*}{1.0297} & 0.99252                                                              \\
\multicolumn{1}{c|}{\textbf{}}                               & \multicolumn{1}{c|}{}                                                                                      &                         & \multicolumn{1}{c|}{(-5.04\%)}                                                            &                         & \multicolumn{1}{c|}{(-3.24\%)}                                                            &                         & \multicolumn{1}{c|}{(-3.32\%)}                                                            &                         & \multicolumn{1}{c|}{(-3.21\%)}                                                            &                         & (-3.61\%)                                                            \\
\multicolumn{1}{c|}{\multirow{2}{*}{\textbf{Vizwiz}}}        & \multicolumn{1}{c|}{\multirow{2}{*}{\textbf{*MCS}}}                                                         & \multirow{2}{*}{1.0626} & \multicolumn{1}{c|}{1.0338}                                                               & \multirow{2}{*}{1.0491} & \multicolumn{1}{c|}{1.01004}                                                              & \multirow{2}{*}{1.0509} & \multicolumn{1}{c|}{1.02508}                                                              & \multirow{2}{*}{1.0505} & \multicolumn{1}{c|}{1.01464}                                                              & \multirow{2}{*}{1.0414} & 1.00578                                                              \\
\multicolumn{1}{c|}{}                                        & \multicolumn{1}{c|}{}                                                                                      &                         & \multicolumn{1}{c|}{(-2.71\%)}                                                            &                         & \multicolumn{1}{c|}{(-3.72\%)}                                                            &                         & \multicolumn{1}{c|}{(-2.46\%)}                                                            &                         & \multicolumn{1}{c|}{(-3.41\%)}                                                            &                         & (-3.42\%)                                                            \\
\multicolumn{1}{c|}{\textbf{}}                               & \multicolumn{1}{c|}{\multirow{2}{*}{\textbf{PR-MCS}}}                                                      & \multirow{2}{*}{0.9769} & \multicolumn{1}{c|}{\textbf{0.29444}}                                                     & \multirow{2}{*}{0.9895} & \multicolumn{1}{c|}{\textbf{0.41498}}                                                     & \multirow{2}{*}{1.442}  & \multicolumn{1}{c|}{\textbf{0.61628}}                                                     & \multirow{2}{*}{1.4526} & \multicolumn{1}{c|}{\textbf{0.6257}}                                                      & \multirow{2}{*}{1.4113} & \textbf{0.62244}                                                     \\
\multicolumn{1}{c|}{\textbf{}}                               & \multicolumn{1}{c|}{}                                                                                      &                         & \multicolumn{1}{c|}{\textbf{(-69.86\%)}}                                                  &                         & \multicolumn{1}{c|}{\textbf{(-58.06\%)}}                                                  &                         & \multicolumn{1}{c|}{\textbf{(-57.26\%)}}                                                  &                         & \multicolumn{1}{c|}{\textbf{(-56.93\%)}}                                                  &                         & \textbf{(-55.90\%)}                                                  \\ \hline
\multicolumn{1}{c|}{\textbf{}}                               & \multicolumn{1}{c|}{\multirow{2}{*}{\textbf{$\dagger$MCS}}}                                                         & \multirow{2}{*}{0.7593} & \multicolumn{1}{c|}{0.70312}                                                              & \multirow{2}{*}{1.0708} & \multicolumn{1}{c|}{1.02584}                                                              & \multirow{2}{*}{1.053}  & \multicolumn{1}{c|}{1.0153}                                                               & \multirow{2}{*}{1.0599} & \multicolumn{1}{c|}{1.02048}                                                              & \multirow{2}{*}{1.06}   & 1.02648                                                              \\
\multicolumn{1}{c|}{\textbf{}}                               & \multicolumn{1}{c|}{}                                                                                      &                         & \multicolumn{1}{c|}{(-7.40\%)}                                                            &                         & \multicolumn{1}{c|}{(-4.20\%)}                                                            &                         & \multicolumn{1}{c|}{(-3.58\%)}                                                            &                         & \multicolumn{1}{c|}{(-3.72\%)}                                                            &                         & (-3.16\%)                                                            \\
\multicolumn{1}{c|}{\textbf{}}                               & \multicolumn{1}{c|}{\multirow{2}{*}{\textbf{*MCS}}}                                                         & \multirow{2}{*}{1.0657} & \multicolumn{1}{c|}{1.0372}                                                               & \multirow{2}{*}{1.0803} & \multicolumn{1}{c|}{1.04696}                                                              & \multirow{2}{*}{1.0575} & \multicolumn{1}{c|}{1.02948}                                                              & \multirow{2}{*}{1.0577} & \multicolumn{1}{c|}{1.02972}                                                              & \multirow{2}{*}{0.963}  & 0.98556                                                              \\
\multicolumn{1}{c|}{\multirow{2}{*}{\textbf{M-FineCapEval}}} & \multicolumn{1}{c|}{}                                                                                      &                         & \multicolumn{1}{c|}{(-2.67\%)}                                                            &                         & \multicolumn{1}{c|}{(-3.09\%)}                                                            &                         & \multicolumn{1}{c|}{(-2.65\%)}                                                            &                         & \multicolumn{1}{c|}{(-2.65\%)}                                                            &                         & (2.34\%)                                                             \\
\multicolumn{1}{c|}{}                                        & \multicolumn{1}{c|}{\multirow{2}{*}{\textbf{PR-MCS}}}                                                      & \multirow{2}{*}{1.0429} & \multicolumn{1}{c|}{\textbf{0.55786}}                                                     & \multirow{2}{*}{0.7261} & \multicolumn{1}{c|}{\textbf{0.4611}}                                                      & \multirow{2}{*}{1.3634} & \multicolumn{1}{c|}{\textbf{0.84882}}                                                     & \multirow{2}{*}{1.3821} & \multicolumn{1}{c|}{\textbf{0.94256}}                                                     & \multirow{2}{*}{1.1551} & \textbf{0.6133}                                                      \\
\multicolumn{1}{c|}{\textbf{}}                               & \multicolumn{1}{c|}{}                                                                                      &                         & \multicolumn{1}{c|}{\textbf{(-46.51\%)}}                                                  &                         & \multicolumn{1}{c|}{\textbf{(-36.50\%)}}                                                  &                         & \multicolumn{1}{c|}{\textbf{(-37.74\%)}}                                                  &                         & \multicolumn{1}{c|}{\textbf{(-31.80\%)}}                                                  &                         & \textbf{(-46.91\%)}                                                  \\
\multicolumn{1}{c|}{\textbf{}}                               & \multicolumn{1}{c|}{\multirow{2}{*}{\textbf{\begin{tabular}[c]{@{}c@{}}PR-MCS\\ (Few-shot)\end{tabular}}}} & \multirow{2}{*}{0.7125} & \multicolumn{1}{c|}{\textbf{0.1584}}                                                      & \multirow{2}{*}{0.589}  & \multicolumn{1}{c|}{\textbf{0.17418}}                                                     & \multirow{2}{*}{1.4862} & \multicolumn{1}{c|}{\textbf{0.43914}}                                                     & \multirow{2}{*}{1.4292} & \multicolumn{1}{c|}{\textbf{0.39644}}                                                     & \multirow{2}{*}{1.4136} & \textbf{0.31398}                                                     \\
\multicolumn{1}{c|}{}                                        & \multicolumn{1}{c|}{}                                                                                      &                         & \multicolumn{1}{c|}{\textbf{(-77.77\%)}}                                                  &                         & \multicolumn{1}{c|}{\textbf{(-70.43\%)}}                                                  &                         & \multicolumn{1}{c|}{\textbf{(-70.45\%)}}                                                  &                         & \multicolumn{1}{c|}{\textbf{(-72.26\%)}}                                                  &                         & \textbf{(-77.79\%)}                                                  \\ \hline
                                                             &                                                                                                            &                         &                                                                                           &                         &                                                                                           &                         &                                                                                           &                          &\multicolumn{3}{l}{$\dagger$ : \citet{multilingualCLIP2022} based}                                                                                                                                                                               \\
                                                             &                                                                                                            &                         &                                                                                           &                         &                                                                                           &                         &                                                                                           &                         & \multicolumn{3}{l}{*: Our Multilingual CLIP based}                                                                                                                                                                         
\end{tabular}
}
\caption{Experiement results table. Each values are represented using the average value for each perturbation. For all four datasets, PR-MCS outperforms the baseline performance for all languages, and the performance has further increased after additional~\dataset~fine-tuning in few-shot settings.}
\label{table2}
\end{table*}

\subsection{Experimental Setup}
\noindent \textbf{Fine-tuning set} We use MSCOCO, the dataset most widely used for image captioning, as the fine-tuning set to enhance the perturbation robustness of MCS. We use the training and validation split of the MSCOCO dataset described by \citet{chen2020uniter}. The number of elements in the training set is 414k. Since only English captions exist in MSCOCO, captions are translated into four other languages using the MBART-50~\cite{liu2020multilingual} MT model.\\

\noindent \textbf{Evaluation set} To comprehensively evaluate the perturbation robustness of PR-MCS, we choose four evaluation sets: MSCOCO, VizWiz~\cite{gurari2020captioning}, Flickr8k~\cite{anitha2019automated}, and M-FineCapEval. As in the fine-tuning, MSCOCO, VizWiz, and Flickr8k, which have only English captions, are translated using the MBART-50 MT model. The evaluation dataset sizes are unified to 3k.\\

\noindent \textbf{Baseline metric} As the baseline of the experiment, we use two MCS metrics. As mentioned above, the MCS metric is configured using CLIP’s visual and multilingual text encoder. The first baseline is the MCS metric constructed using the multilingual CLIP text encoder implemented by \citet{multilingualCLIP2022} as the backbone. The second baseline is the MCS metric constructed using the multilingual text encoder trained by the teacher learning method described in Section~\ref{MCS}.

\subsection{Perturbation configuration} 
We select the following five criteria to perturb the sentences in the fine-tuning and evaluation sets. The criteria below are error types commonly found in model-generated captions. These criteria are part of the checklist proposed by \citet{sai2021perturbation}. We select five orthogonal criteria. Each perturbation example is shown in Figure~\ref{fig5}.\\

\noindent \textbf{Repetition} Repeated words are found in several model-generated captions. A well-known problem is that the transformer model is vulnerable because it does not capture repetitive perturbation well at the embedding level. We give each word token a repeating perturbation with a probability of 0.4.\\

\noindent \textbf{Removal} Among the sentences given a low score in the evaluation dataset for the image captioning metric, such as Composite~\cite{aditya2015images} or Pascal50s~\cite{vedantam2015cider}, some word tokens are removed, and incomplete sentences are found. We configure perturbation by removing some tokens to reflect this noise. Each word token is drawn with a probability of 0.4.\\

\noindent \textbf{Masking} Masking is a perturbation in which randomly selected tokens in the caption are replaced with [Mask] tokens. When lexical noise is given in units of tokens, the meaning of the corresponding token disappears, but unlike in the Removal case, the position is maintained. Position information can be critical in a reference-free metric based on a transformer model such as CLIPScore~\cite{dai2019transformer, devlin2018bert, ramachandran2019stand, wu2021rethinking}. Therefore, even if the [Mask] token does not appear in the generated caption, we select Masking perturbation as the criterion, separate from Removal, to address the above case. Each word token is replaced with a [Mask] token with a probability of 0.4.\\

\noindent \textbf{Jumble} We generate perturbed samples using random-order permutation at the token level in the original reference caption. The model composing the metric can see all tokens of the sentence, including visual content, but considerable noise is introduced into the position information.\\

\noindent \textbf{Substitution} Substitution involves changing the positions of key elements in a sentence. In the case of \dataset, substitution is performed using critical objects annotated by human experts. In the remaining three datasets, nouns in the caption are extracted, and their positions are changed. The perturbed caption includes all elements that exist in the original caption, but unlike in the Jumble case, it does not deform the grammatical structure at all.
Detecting substitution noise well is the most challenging task because it requires judging semantic correspondence to visual content perfectly.

\subsection{Perturbation robustness evaluation}
\label{results}
We report the main results for all datasets and languages in Table~\ref{table2} and Figure~\ref{fig6}. The robust evaluation metric is expected to give lower scores to perturbed captions than to original captions.

Each graph of Figure~\ref{fig6} shows the experimental results for MSCOCO, VizWiz, Flickr8k, and M-FineCapEval by perturbation. Each point represents the average results for five languages for one perturbation. It shows how much score drop the perturbed caption has from the original caption. The green line indicates PR-MCS, and the blue and red lines refer to the two baseline multilingual CLIPScores. The scores of the perturbed caption by baseline metrics do not differ much from those of the original caption for any perturbation methods. In some cases, the scores for the perturbed captions are higher than those for the original captions.

However, our metric exhibits a significant score decrease for all perturbations compared to the original captions, which means that the metric can clearly distinguish when the perturbation is applied. In other words, our metric exhibits robustness for all perturbations in the evaluation dataset. 
In particular, even in the cases of Repetition and Substitution, which are known to be challenging perturbations, PR-MCS detects perturbations very well, while baseline metrics do not capture perturbations at all.

Table~\ref{table2} shows the score of the original caption and the average score of the perturbed caption given by the metrics for each of the four evaluation sets for each language. The result shows how much the percentage of the score decreased for each perturbation in comparison to the original caption. The results for the baseline metrics show that the score decrease for the average perturbation is very small, i.e., approximately 3\%, relative to the original caption. 
It is difficult to say that the metric can distinguish the perturbed caption from the original caption based on such a slight difference.
In contrast, in the case of PR-MCS, the percentage decrease for the perturbed caption ranges from 50\% to 70\%. Clearly, our proposed method exhibits perturbation robustness in the metric score and can identify perturbed captions through anomaly detection only with a performance drop. In the cases of Vizwiz, Flickr8k, and M-FineCapEval, the performance is outstanding even though the distributions are not trained in fine-tuning.

\subsection{Few-shot setting for M-FineCapEval}
As the results summarized in section~\ref{results} show, PR-MCS is much more robust to perturbation than the baselines in \dataset. However, the performance degradation for perturbed captions in \dataset is lower than for MSCOCO, VizWiz, and Flickr8k (e.g., -55.66\% to -39.89\% in average from MSCOCO). We attribute this to the distribution shift from the sequence length difference between the MSCOCO fine-tuning set and the \dataset test set. VizWiz and Flickr8k are composed of short captions, so there is not much difference in caption length between them and MSCOCO. Therefore, to check whether the distribution can be learned when some information about the \dataset 3K test set is provided, we perform additional experiments on \dataset with a few-shot setting. We split \dataset into subsets proportioned 1:9 in size and use only 300 perturbed captions as the few-shot input.

The experimental results for the few-shot setting are shown in Table~\ref{table2} and Figure~\ref{fig6} (the yellow line in rightmost graph). When the distribution for the fine-grained caption is given, the overall performance in perturbation detection, in terms of the average score, increases for all five languages. These results show that lexical noise in long sentences is more reliably captured by learning a small number of samples with a few-shot setting. The experimental results for all languages and all perturbations for each dataset are provided in the Appendix~\ref{appendix:all_results_table}.

\begin{table}[t]
\renewcommand{\arraystretch}{1.5} 
\centering
\resizebox{\columnwidth}{!}{%
\begin{tabular}{c|cc|cc|c|c}
\hline
\multirow{2}{*}{}  & \multicolumn{2}{c|}{\textbf{Flickr8k\_Exp}} & \multicolumn{2}{c|}{\textbf{CapEval1k}} & \multirow{2}{*}{\textbf{\begin{tabular}[c]{@{}c@{}} Multi- \\ Lingual\end{tabular}}} & \multirow{2}{*}{\textbf{\begin{tabular}[c]{@{}c@{}} Perturbation \\ Robustness \end{tabular}}} \\ \cline{2-5}
                   & \textbf{$\tau_{c}$}       & \textbf{$\rho$}      & \textbf{$\tau_{c}$}       & \textbf{$\rho$}       &                                                                                              \\ \hline
\textbf{CLIPScore} & 0.515       & 0.715           & 0.27                 & 0.408            & \xmark    & \xmark                                                                                        \\ \hline
\textbf{MCS}       & 0.439                & 0.599           & 0.253       & 0.389            & \cmark    & \xmark                                                                                       \\ \hline
\textbf{PR-MCS}    & 0.506       & 0.656           & 0.246       & 0.394            & \cmark  & \cmark                                                                                   \\ \hline
\end{tabular}%
}
\caption{Flickr8k\_Expert and CapEval1k correlations with human judgment.}
\label{table3}
\end{table}

\subsection{Correlations with human judgement}
Table~\ref{table3} shows that PR-MCS is an useful image captioning metric with high correlation with human judgment. 
Flickr8k\_Expert \cite{hodosh2013framing} and CapEval1k \cite{lee2021umic} are evaluation sets for measuring the performance of image captioning metric, and the higher the Kendall tau-c ($\tau_{c}$) value \cite{kendall1938new} and the Pearson correlation coefficient ($\rho$)~\cite{benesty2009pearson}, indicators for viewing the correlation with human judgment, the better. The Kendall tau-c value is the similarity between the two variables based on ranking, and the Pearson correlation coefficient is a measure of linear correlation between two sets of data.

We conduct experiments on the CLIPScore only in English, and for MCS and PR-MCS, we translate the two datasets into four languages and then report the average value. Due to the incompleteness and bias of the MT model, MCS shows slightly lower correlations with human evaluation compared to the CLIPScore. On the other hand, PR-MCS shows a higher correlations than MCS, or even similar to the CLIPScore, even though it has been granted perturbation robustness as shown in Section~\ref{results}. From these results, it can be seen that PR-MCS is a metric highly correlated with human evaluation while capturing perturbation well.

\section{Conclusion}
\label{sec:conclusion}
In this paper, we propose PR-MCS, a perturbation-robust metric for multilingual image captioning using language-agnostic fine-tuning. PR-MCS, developed by fine-tuning the text encoder of CLIP, can distinguish lexically perturbed text from original text. We also propose a new fine-grained multilingual evaluation set, \dataset, for use in perturbation robustness evaluation. Experimental results for existing datasets and our new dataset show that PR-MCS detects perturbation well and is robust to perturbation in multiple languages.  

\clearpage

\section*{Limitations}
\label{sec:limitations}
\noindent \textbf{Model bias of machine translation in training}

\noindent In our study, an evaluation set is created by directly annotating languages other than English to remove the bias of the machine translation (MT) model in the evaluation phase.
However, in the training phase, the dataset size is too large to annotate directly in multiple languages other than English.
Therefore, the pre-training set and the fine-tuning set are translated into other languages by utilizing the MT model, so we have no choice but to depend on the performance of the MT model and avoid model bias.

A better model can be obtained if training is carried out using a large corpus with captions annotated in multiple languages corresponding to one image.
So far, such a multilingual multimodal large corpus is insufficient.
Recently, several large corpora having image-text pairs in various languages, such as LAION\_5B \cite{schuhmann2022laion}, have been introduced, but such sets have a structure in which text is matched in one language among several languages on one image.
In future work, we plan to extend our research by using these datasets to consider images as universal language representations.

\bibliography{acl}
\bibliographystyle{acl_natbib}

\clearpage
\appendix
\label{sec:appendix}

\section{Pre-training Details}
Like \citet{multilingualCLIP2022}'s method, our multilingual CLIP is trained by pre-training through teacher learning using MSE loss as shown on the left of the Figure \ref{fig2}. The datasets used for pre-training are GCC \cite{wang2019learning}, VizWiz, and MSCOCO with total 2.2M sentences. Each English sentence is translated into German, Spanish, French, and Japanese using MBART-50. The pre-trained CLIP model used as the teacher model is the RN50X4 model, and the Distill-Multilingual BERT \cite{sanh2019distilbert} is used as the student text encoder. The model is trained in a total of 5 epochs, and it takes about 20 hours per epoch with our computing power . 
\label{pretraining_details}

\section{Experimental Details}
\subsection{Reproductabilty checklists} 
\noindent \textbf{Dataset and Source code} We provide our pre-training, fine-tuning, and evaluation source code along with configuration code for perturbations as supplementary materials.
We will publicly release our dataset~\dataset, and the full codes with weight parameters. \\

\noindent \textbf{Computing Resources} AMD Ryzen Threadripper 2950X (3.50 GHz) with GeForce GTX 2080 Ti is used for the experiments. All codes are implemented on Python 3.6.15 and PyTorch 1.7.1. The fine-tuning of each model trains 5 epochs, and takes about 6 hours per epoch.\\

\noindent \textbf{Number of Parameters} The number of parameter of our multilingual CLIP is about 66M as like as Distill-Multilingual BERT. \\

\noindent \textbf{Train-Valid-Test split} MSCOCO used for fine-tuning consists of 414k training set and 25k validation set. We split the training set by 9:1 and used it for fine-tuning and validation. We also randomly extracted 3k samples from the existing validation set and used it as a test set.

\subsection{Hyper-parameters}
\noindent \textbf{Hyper-parameters for fine-tuning} In order to find the best-performing model, we conducted an experiment on 16 hyper-parameter combinations ($\lambda_{1}:0\sim 0.5, \lambda_{2}:0\sim0.1, \lambda_{3}:0\sim0.1$). 
The hyper-parameter was manually tuned based on the effective detection of lexical noise while maintaining high human correlation, and finally, the best-performing $\lambda$ values of the objective function for fine-tuning are as follows:
$\lambda_{1}=0.1, \lambda_{2}=0.05, \lambda_{3}=0.05$. \\ 

\noindent \textbf{Hyper-parameters for optimizer} We use AdamW \cite{loshchilov2017decoupled} optimizer with $\beta_{1}=0.9,\beta_{2}=0.999, \epsilon=1e-8$. The initial learning rate is $5e-5$.

\section{\dataset eval set examples}
\begin{figure}[h]
\centering
\includegraphics[width=1.0\columnwidth]{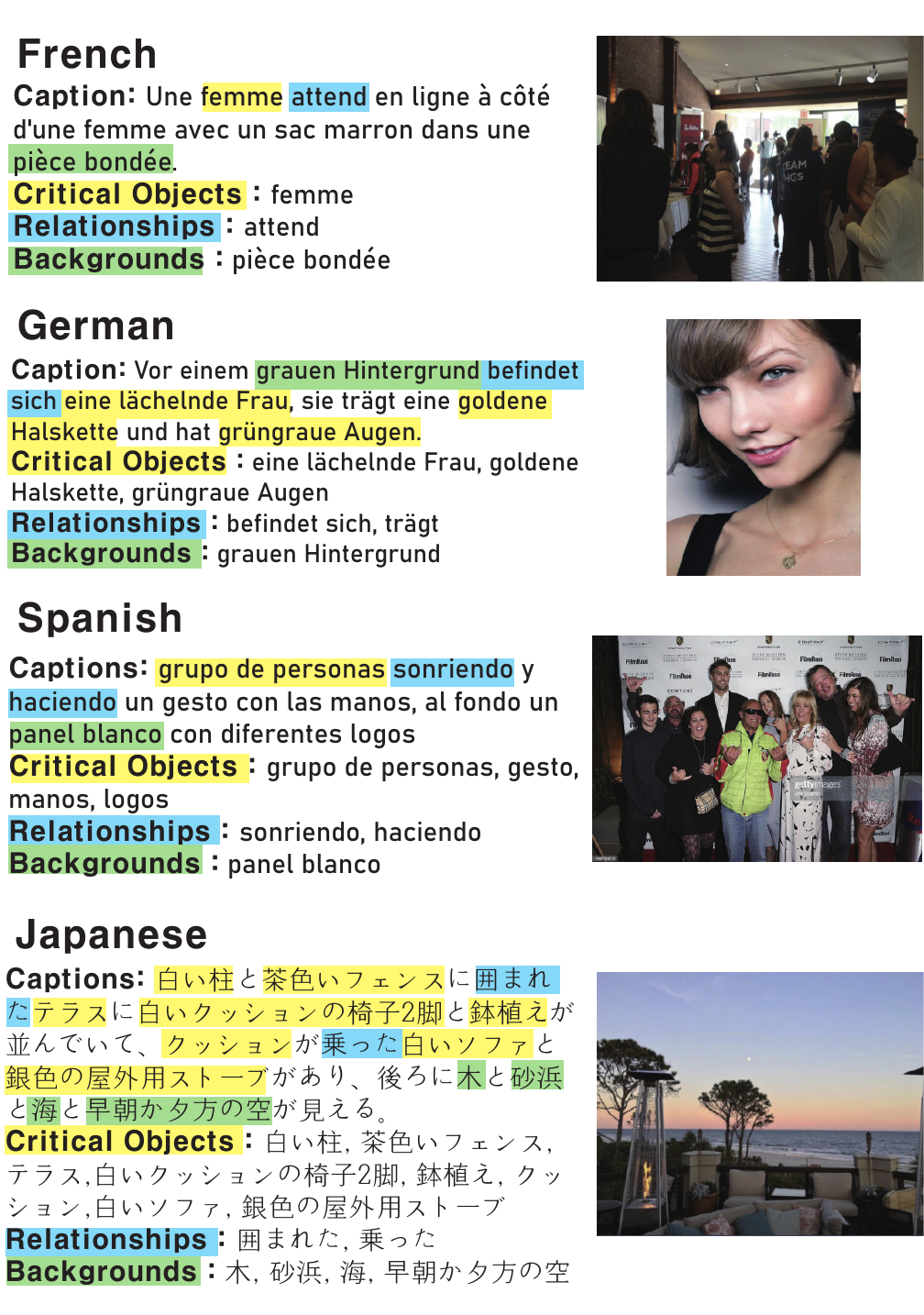} 
\caption{\dataset~eval set examples for each languages.}
\label{fig7}
\end{figure}

The examples of the \dataset eval set for languages other than English can be seen in Figure~\ref{fig7}.



\section{Perturbed caption examples}
\begin{figure}[t]
\centering
\includegraphics[width=1.0\columnwidth]{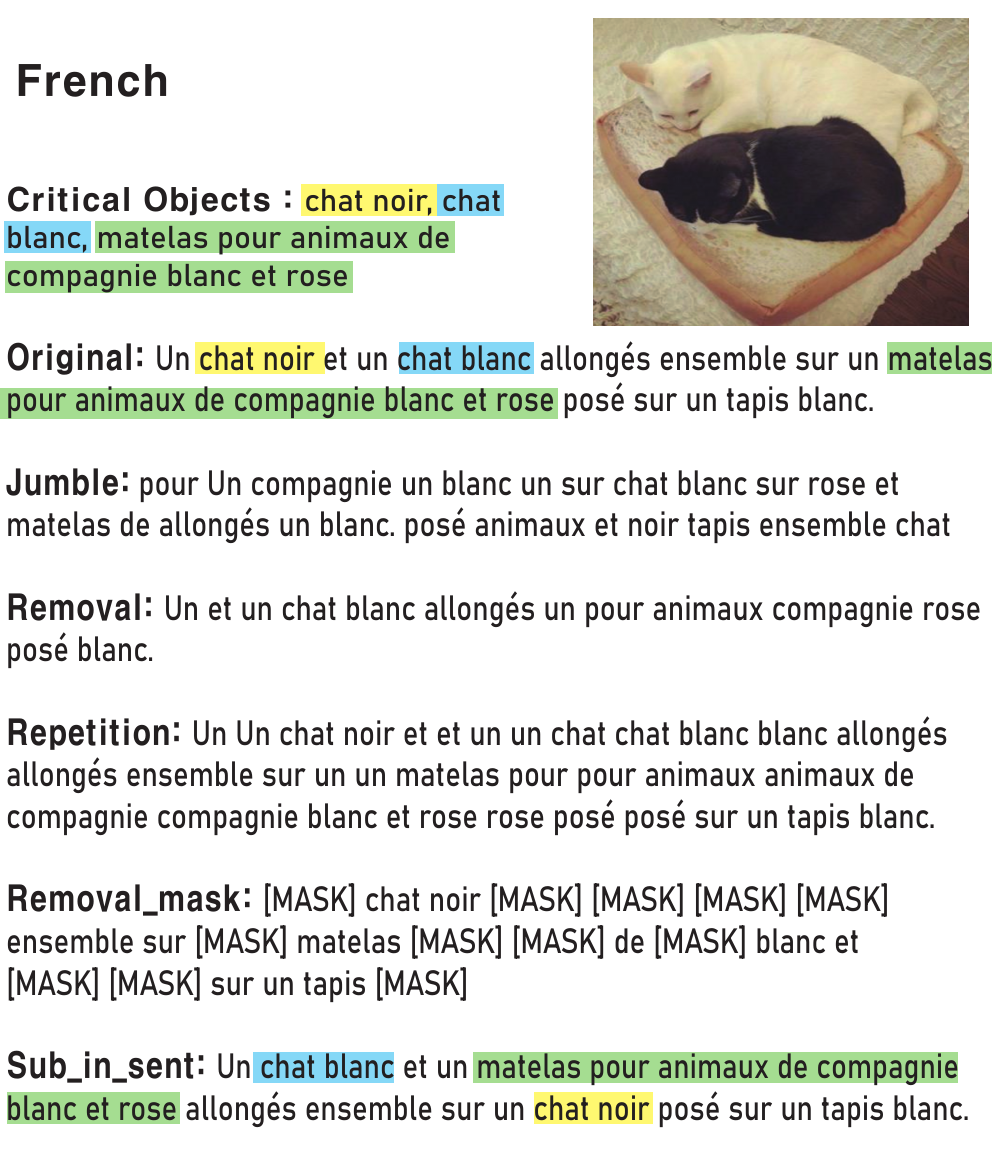} 
\caption{Eval set perturbed captions example (FR).}
\label{fig8}
\end{figure}

\begin{figure}[t]
\centering
\includegraphics[width=1.0\columnwidth]{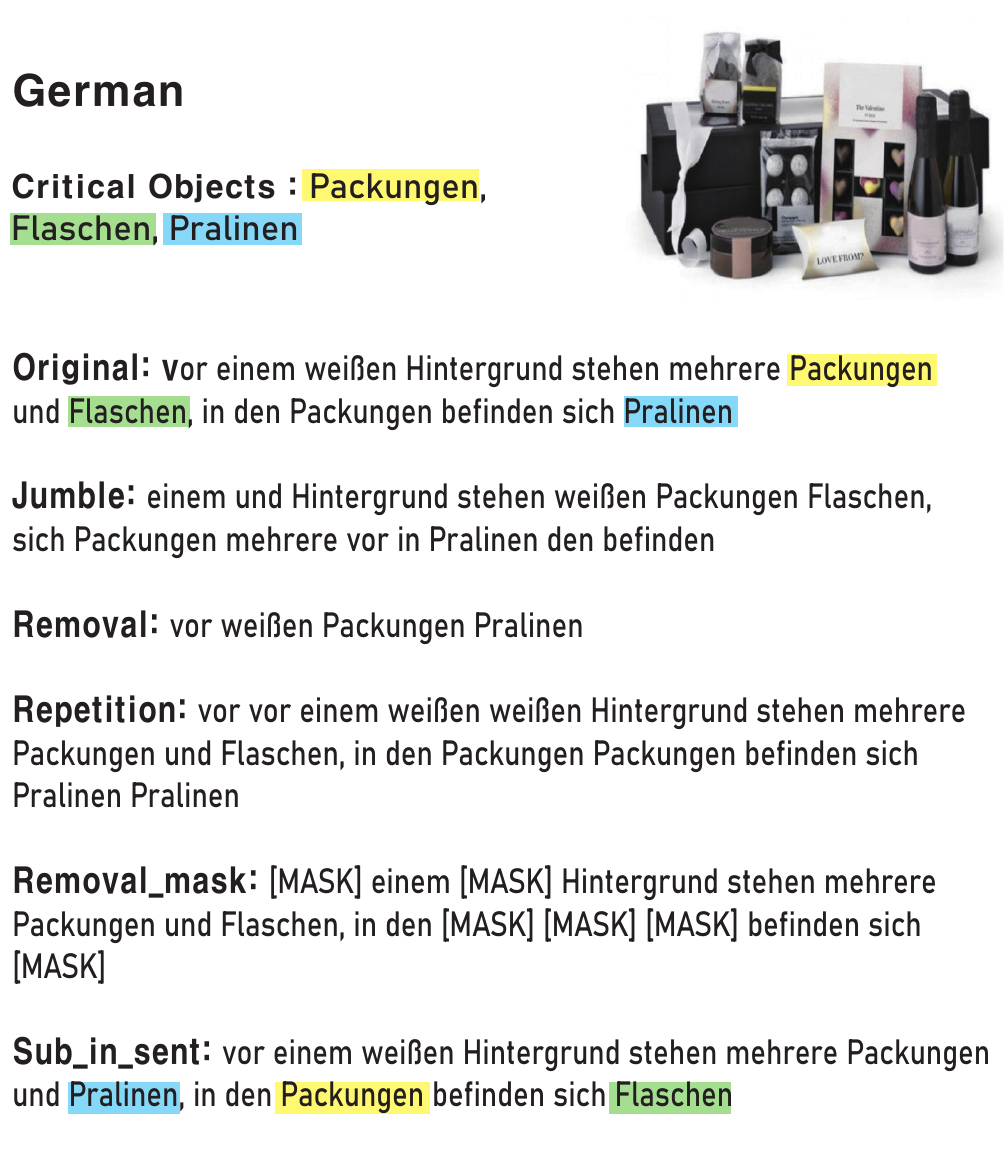} 
\caption{Eval set perturbed captions example (DE).}
\label{fig9}
\end{figure}

\begin{figure}[t]
\centering
\includegraphics[width=1.0\columnwidth]{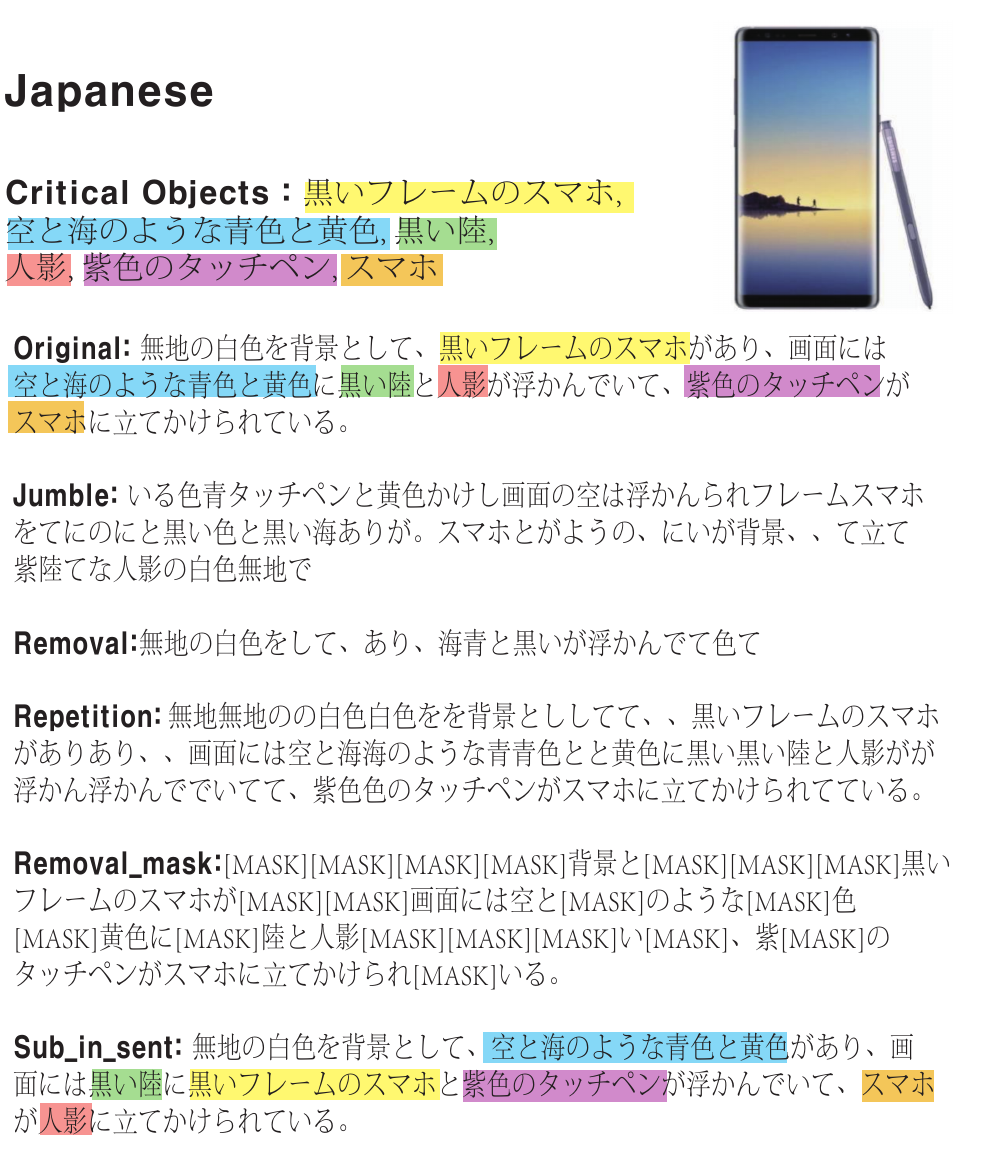} 
\caption{Eval set perturbed captions example (JA).}
\label{fig10}
\end{figure}

\begin{figure}[t]
\centering
\includegraphics[width=1.0\columnwidth]{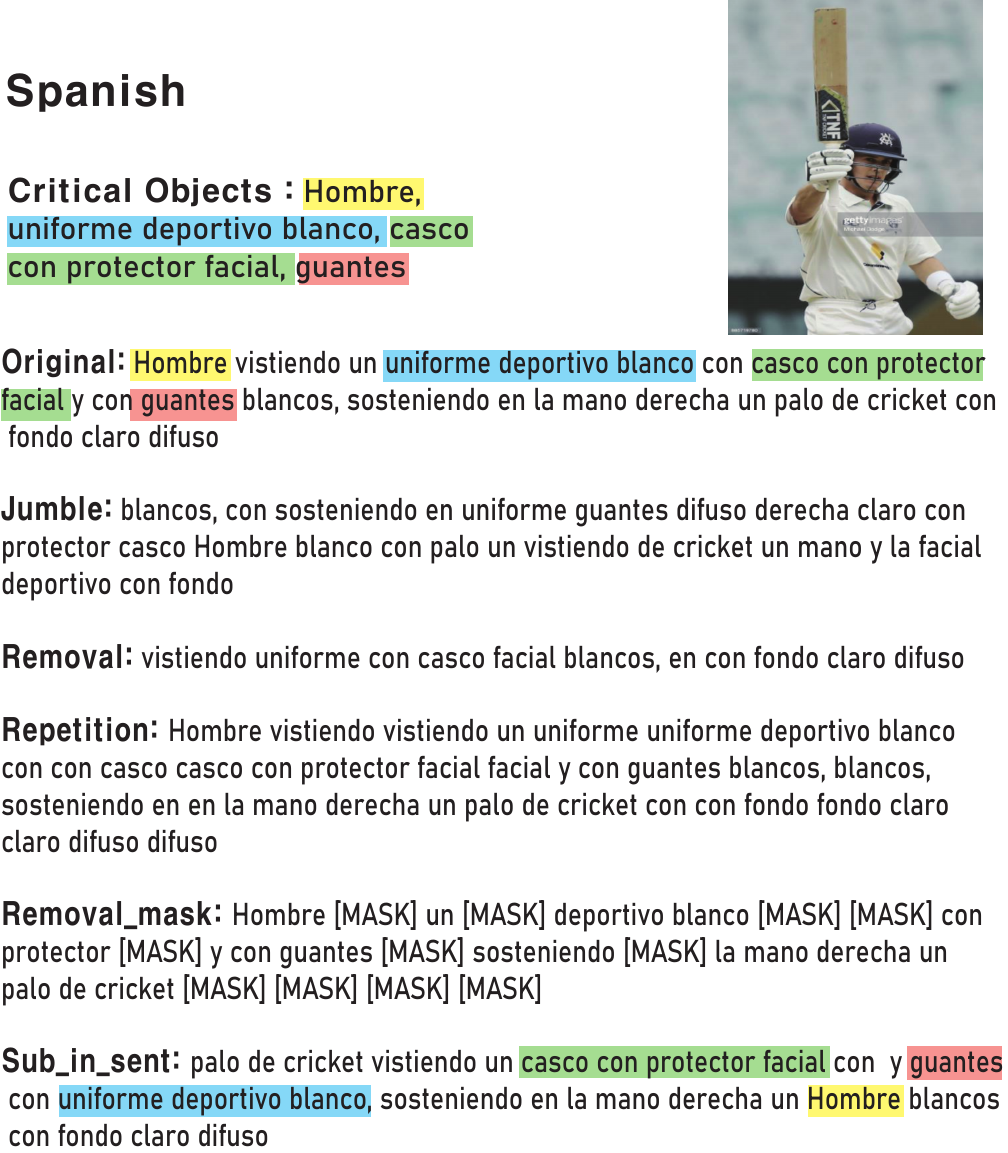} 
\caption{Eval set perturbed captions example (ES).}
\label{fig11}
\end{figure}

The examples of the perturbed captions for languages other than English can be seen in Figure~\ref{fig8}-Figure~\ref{fig11}.
The critical objects shuffled for in-sentence substitution perturbation are displayed using each color.

\section{Implementation Details}
In Alg.~\ref{alg1}, we show the Python implementation of each perturbation criterion: \textit{"Repetition"}, \textit{"Removal"}, \textit{"Masking"}, \textit{"Jumble"}, and \textit{"Substitution"}.

\section{All results tables}
\label{appendix:all_results_table}

\noindent \textbf{MSCOCO}
The results for all perturbation of all languages for MSCOCO 3k eval set can be found in Table \ref{table4}.\\
\noindent \textbf{Flickr8k}
The results for all perturbation of all languages for Flickr8k eval set can be found in Table \ref{table5}.\\
\noindent \textbf{VizWiz}
The results for all perturbation of all languages for Vizwiz eval set can be found in Table \ref{table6}.\\
\noindent \textbf{\dataset}
The results for all perturbation of all languages for~\dataset eval set can be found in Table \ref{table7}.

\begin{algorithm*}
\caption{Python implementation of perturbation}\label{alg1} 
\vspace{2mm}

\lstset{language=Python}
\lstset{label={lst:code_direct}}
\lstset{basicstyle=\footnotesize}
\begin{lstlisting}
def rep_rem_mask(caption_list): # Repetition, Removal, and Masking
    caption_rp = []
    caption_rm = []
    caption_rm_mask = []

    for i in range(len(caption_list)):
        words = caption_list[i].split()
        substitued_rp = []
        substitued_rm = []
        substitued_rm_mask = []
        substitued_rmrp = []
        for j in range(len(words)):
            substitued_rp.append(words[j])
            substitued_rm_mask.append(words[j])
            if random.random() > threshold:
                substitued_rp.append(words[j])
                substitued_rm.append(words[j])
                substitued_rm_mask[-1] = '[MASK]'
            elif random.random() > threshold:
                substitued_rmrp.append(words[j])  
        caption_rp.append(" ".join(substitued_rp))
        caption_rm.append(" ".join(substitued_rm))
        caption_rm_mask.append(" ".join(substitued_rm_mask))
        
    return caption_rp, caption_rm, caption_rm_mask

def jumble(caption_list): # Jumble
    caption_jumble = []
    for i in range(len(caption_list)):
        words = caption_list[i].split()
        random.shuffle(words)
        caption_jumble.append(" ".join(words))
    return caption_jumble
    
def sub_in_sent(caption_list, critical_obj_list): # In-sentence substitution
    caption_sub_in = []
    for i in range(len(caption_list)):
        current_caption = caption_list[i]
        current_critical_obj_list = critical_obj_list[i]
        shuffled = current_critical_obj_list.copy()
        words = caption_list[i].split()
        
        if len(current_critical_obj_list) < 2:
            caption_sub_in.append(" ".join(words))
        else :
            while current_critical_obj_list == shuffled:
                random.shuffle(shuffled)
            
            target = current_caption
            for j in range(len(shuffled)):
                target = shuffled[j].join(target.rsplit(current_critical_obj_list[j],1))
            caption_sub_in.append(target)
            
    return caption_sub_in

\end{lstlisting}

\end{algorithm*}

\clearpage

\clearpage
\begin{table*}[p]
\renewcommand{\arraystretch}{1.3}
\centering
\resizebox{\textwidth}{!}{
\begin{tabular}{ccccccccc}
\hline
\multicolumn{1}{c|}{\textbf{Language}}                  & \multicolumn{1}{c|}{\textbf{Metric}}                        & \textbf{Original}    & \textbf{Repetition}                                         & \textbf{Removal}                                            & \textbf{\begin{tabular}[c]{@{}c@{}}Masking\end{tabular}} & \textbf{Jumble}                                             & \textbf{Substitution}                                             & \textbf{Average}                                                      \\ \hline
\multicolumn{1}{c|}{\textbf{}}         & \multicolumn{1}{c|}{\textbf{$\dagger$MCS}}    & 0.7944               & \begin{tabular}[c]{@{}c@{}}0.7871\\ (-0.92\%)\end{tabular}  & \begin{tabular}[c]{@{}c@{}}0.7241\\ (-8.84\%)\end{tabular}  & \begin{tabular}[c]{@{}c@{}}0.7137\\ (-10.15\%)\end{tabular}           & \begin{tabular}[c]{@{}c@{}}0.7573\\ (-4.67\%)\end{tabular}  & \begin{tabular}[c]{@{}c@{}}0.7817\\ (-1.59\%)\end{tabular}  & \begin{tabular}[c]{@{}c@{}}0.7442\\ (-5.23\%)\end{tabular}            \\
\multicolumn{1}{c|}{\textbf{English}}  & \multicolumn{1}{c|}{\textbf{*MCS}}    & 1.0939               & \begin{tabular}[c]{@{}c@{}}1.0963\\ (0.21\%)\end{tabular}   & \begin{tabular}[c]{@{}c@{}}1.0177\\ (-6.96\%)\end{tabular}  & \begin{tabular}[c]{@{}c@{}}0.987\\ (-9.77\%)\end{tabular}             & \begin{tabular}[c]{@{}c@{}}1.0966\\ (-0.24\%)\end{tabular}  & \begin{tabular}[c]{@{}c@{}}1.0928\\ (-0.10\%)\end{tabular}  & \begin{tabular}[c]{@{}c@{}}1.05808\\ (-3.27\%)\end{tabular}           \\
\multicolumn{1}{c|}{\textbf{}}         & \multicolumn{1}{c|}{\textbf{PR-MCS}} & 1.4177               & \begin{tabular}[c]{@{}c@{}}0.8800\\ (-37.93\%)\end{tabular} & \begin{tabular}[c]{@{}c@{}}0.3103\\ (-78.11\%)\end{tabular} & \begin{tabular}[c]{@{}c@{}}0.0112\\ (-99.21\%)\end{tabular}           & \begin{tabular}[c]{@{}c@{}}0.0433\\ (-96.95\%)\end{tabular} & \begin{tabular}[c]{@{}c@{}}0.2534\\ (-82.13\%)\end{tabular} & \textbf{\begin{tabular}[c]{@{}c@{}}0.29964\\ (-78.86\%)\end{tabular}} \\ \hline
\multicolumn{1}{c|}{\textbf{}}         & \multicolumn{1}{c|}{\textbf{$\dagger$MCS}}    & 1.0743               & \begin{tabular}[c]{@{}c@{}}1.0764\\ (0.19\%)\end{tabular}   & \begin{tabular}[c]{@{}c@{}}0.9989\\ (-7.01\%)\end{tabular}  & \begin{tabular}[c]{@{}c@{}}0.9501\\ (-11.56\%)\end{tabular}           & \begin{tabular}[c]{@{}c@{}}1.0792\\ (0.45\%)\end{tabular}   & \begin{tabular}[c]{@{}c@{}}1.0735\\ (-0.07\%)\end{tabular}  & \begin{tabular}[c]{@{}c@{}}1.03562\\ (-3.60\%)\end{tabular}           \\
\multicolumn{1}{c|}{\textbf{German}}   & \multicolumn{1}{c|}{\textbf{*MCS}}    & 1.0839               & \begin{tabular}[c]{@{}c@{}}1.0851\\ (-0.11\%)\end{tabular}  & \begin{tabular}[c]{@{}c@{}}1.0119\\ (-6.64\%)\end{tabular}  & \begin{tabular}[c]{@{}c@{}}1.0044\\ (-7.33\%)\end{tabular}            & \begin{tabular}[c]{@{}c@{}}1.0887\\ (0.44\%)\end{tabular}   & \begin{tabular}[c]{@{}c@{}}1.0855\\ (0.15\%)\end{tabular}   & \begin{tabular}[c]{@{}c@{}}1.05512\\ (-2.66\%)\end{tabular}           \\
\multicolumn{1}{c|}{\textbf{}}         & \multicolumn{1}{c|}{\textbf{PR-MCS}} & 1.3153               & \begin{tabular}[c]{@{}c@{}}0.9082\\ (-30.95\%)\end{tabular} & \begin{tabular}[c]{@{}c@{}}0.686\\ (-47.84\%)\end{tabular}  & \begin{tabular}[c]{@{}c@{}}0.0175\\ (-98.67\%)\end{tabular}           & \begin{tabular}[c]{@{}c@{}}0.166\\ (-87.38\%)\end{tabular}  & \begin{tabular}[c]{@{}c@{}}0.8004\\ (-39.15\%)\end{tabular} & \textbf{\begin{tabular}[c]{@{}c@{}}0.51562\\ (-60.80\%)\end{tabular}} \\ \hline
\multicolumn{1}{c|}{\textbf{}}         & \multicolumn{1}{c|}{\textbf{$\dagger$MCS}}    & 1.0719               & \begin{tabular}[c]{@{}c@{}}1.0753\\ (0.32\%)\end{tabular}   & \begin{tabular}[c]{@{}c@{}}0.9983\\ (-6.87\%)\end{tabular}  & \begin{tabular}[c]{@{}c@{}}0.9453\\ (-11.81\%)\end{tabular}           & \begin{tabular}[c]{@{}c@{}}1.0768\\ (0.46\%)\end{tabular}   & \begin{tabular}[c]{@{}c@{}}1.0692\\ (-0.25\%)\end{tabular}  & \begin{tabular}[c]{@{}c@{}}1.03298\\ (-3.63\%)\end{tabular}           \\
\multicolumn{1}{c|}{\textbf{French}}   & \multicolumn{1}{c|}{\textbf{*MCS}}    & 1.0837               & \begin{tabular}[c]{@{}c@{}}1.0845\\ (0.07\%)\end{tabular}   & \begin{tabular}[c]{@{}c@{}}1.0152\\ (-6.32\%)\end{tabular}  & \begin{tabular}[c]{@{}c@{}}0.9918\\ (-8.48\%)\end{tabular}            & \begin{tabular}[c]{@{}c@{}}1.0878\\ (0.37\%)\end{tabular}   & \begin{tabular}[c]{@{}c@{}}1.0857\\ (0.18\%)\end{tabular}   & \begin{tabular}[c]{@{}c@{}}1.053\\ (-2.83\%)\end{tabular}             \\
\multicolumn{1}{c|}{\textbf{}}         & \multicolumn{1}{c|}{\textbf{PR-MCS}} & 1.4127               & \begin{tabular}[c]{@{}c@{}}1.0021\\ (-29.21\%)\end{tabular} & \begin{tabular}[c]{@{}c@{}}0.9086\\ (-35.68\%)\end{tabular} & \begin{tabular}[c]{@{}c@{}}0.0091\\ (-99.36\%)\end{tabular}           & \begin{tabular}[c]{@{}c@{}}0.2873\\ (-76.66\%)\end{tabular} & \begin{tabular}[c]{@{}c@{}}1.1167\\ (-20.95\%)\end{tabular} & \textbf{\begin{tabular}[c]{@{}c@{}}0.66434\\ (-52.97\%)\end{tabular}} \\ \hline
\multicolumn{1}{c|}{\textbf{}}         & \multicolumn{1}{c|}{\textbf{$\dagger$MCS}}    & 1.0564               & \begin{tabular}[c]{@{}c@{}}1.0591\\ (0.26\%)\end{tabular}   & \begin{tabular}[c]{@{}c@{}}0.9865\\ (-6.61\%)\end{tabular}  & \begin{tabular}[c]{@{}c@{}}0.9331\\ (-11.67\%)\end{tabular}           & \begin{tabular}[c]{@{}c@{}}1.0624\\ (0.57\%)\end{tabular}   & \begin{tabular}[c]{@{}c@{}}1.0528\\ (-0.34\%)\end{tabular}  & \begin{tabular}[c]{@{}c@{}}1.01878\\ (-3.56\%)\end{tabular}           \\
\multicolumn{1}{c|}{\textbf{Spanish}}  & \multicolumn{1}{c|}{\textbf{*MCS}}    & 1.0799               & \begin{tabular}[c]{@{}c@{}}1.0807\\ (0.07\%)\end{tabular}   & \begin{tabular}[c]{@{}c@{}}1.012\\ (-6.29\%)\end{tabular}   & \begin{tabular}[c]{@{}c@{}}0.9909\\ (-8.24\%)\end{tabular}            & \begin{tabular}[c]{@{}c@{}}1.0846\\ (0.44\%)\end{tabular}   & \begin{tabular}[c]{@{}c@{}}1.0812\\ (0.12\%)\end{tabular}   & \begin{tabular}[c]{@{}c@{}}1.04988\\ (-2.77\%)\end{tabular}           \\
\multicolumn{1}{c|}{\textbf{}}         & \multicolumn{1}{c|}{\textbf{PR-MCS}} & 1.4086               & \begin{tabular}[c]{@{}c@{}}1.0316\\ (-26.76\%)\end{tabular} & \begin{tabular}[c]{@{}c@{}}0.8841\\ (-37.24\%)\end{tabular} & \begin{tabular}[c]{@{}c@{}}0.0101\\ (-99.28\%)\end{tabular}           & \begin{tabular}[c]{@{}c@{}}0.2882\\ (-79.54\%)\end{tabular} & \begin{tabular}[c]{@{}c@{}}1.193\\ (-15.31\%)\end{tabular}  & \textbf{\begin{tabular}[c]{@{}c@{}}0.6814\\ (-51.63\%)\end{tabular}}  \\ \hline
\multicolumn{1}{c|}{\textbf{}}         & \multicolumn{1}{c|}{\textbf{$\dagger$MCS}}    & 1.0727               & \begin{tabular}[c]{@{}c@{}}1.0769\\ (0.39\%)\end{tabular}   & \begin{tabular}[c]{@{}c@{}}1.0001\\ (-6.68\%)\end{tabular}  & \begin{tabular}[c]{@{}c@{}}0.9247\\ (-13.80\%)\end{tabular}           & \begin{tabular}[c]{@{}c@{}}1.0819\\ (0.86\%)\end{tabular}   & \begin{tabular}[c]{@{}c@{}}1.0726\\ (0.01\%)\end{tabular}   & \begin{tabular}[c]{@{}c@{}}1.03122\\ (-3.85\%)\end{tabular}           \\
\multicolumn{1}{c|}{\textbf{Japanese}} & \multicolumn{1}{c|}{\textbf{*MCS}}    & 1.0286               & \begin{tabular}[c]{@{}c@{}}1.0348\\ (0.60\%)\end{tabular}   & \begin{tabular}[c]{@{}c@{}}0.9669\\ (-6.00\%)\end{tabular}  & \begin{tabular}[c]{@{}c@{}}0.9105\\ (-11.48\%)\end{tabular}           & \begin{tabular}[c]{@{}c@{}}1.0294\\ (0.08\%)\end{tabular}   & \begin{tabular}[c]{@{}c@{}}1.0271\\ (-0.15\%)\end{tabular}  & \begin{tabular}[c]{@{}c@{}}0.9813\\ (-3.39\%)\end{tabular}            \\
\multicolumn{1}{c|}{\textbf{}}         & \multicolumn{1}{c|}{\textbf{PR-MCS}} & 1.3948               & \begin{tabular}[c]{@{}c@{}}1.3114\\ (-5.97\%)\end{tabular}  & \begin{tabular}[c]{@{}c@{}}1.0539\\ (-24.44\%)\end{tabular} & \begin{tabular}[c]{@{}c@{}}0.0116\\ (-99.16\%)\end{tabular}           & \begin{tabular}[c]{@{}c@{}}0.9114\\ (-34.66\%)\end{tabular} & \begin{tabular}[c]{@{}c@{}}1.3128\\ (-5.88\%)\end{tabular}  & \textbf{\begin{tabular}[c]{@{}c@{}}0.92022\\ (-34.02\%)\end{tabular}} \\ \hline
\multicolumn{1}{l}{}                   & \multicolumn{1}{l}{}                 & \multicolumn{1}{l}{} & \multicolumn{1}{l}{}                                        & \multicolumn{1}{l}{}                                        & \multicolumn{1}{l}{}                                                  & \multicolumn{3}{l}{$\dagger$ : \citet{multilingualCLIP2022} based}                                                                                                                                                                 \\
\multicolumn{1}{l}{}                   & \multicolumn{1}{l}{}                 & \multicolumn{1}{l}{} & \multicolumn{1}{l}{}                                        & \multicolumn{1}{l}{}                                        & \multicolumn{1}{l}{}                                                  & \multicolumn{3}{l}{*: Our Multilingual CLIP based}                                                                                                                                               
\end{tabular}
}
\caption{MSCOCO 3k results table.}
\label{table4}
\end{table*}
\begin{table*}[p]
\renewcommand{\arraystretch}{1.3}
\centering
\resizebox{\textwidth}{!}{
\begin{tabular}{ccccccccc}
\hline
\multicolumn{1}{c|}{\textbf{Language}}                  & \multicolumn{1}{c|}{\textbf{Metric}}                  & \textbf{Original}       & \textbf{Repetition} & \textbf{Removal} & \textbf{\begin{tabular}[c]{@{}c@{}}Masking\end{tabular}} & \textbf{Jumble} & \textbf{Substitution} & \textbf{Average}    \\ \hline
\multicolumn{1}{c|}{}                                   & \multicolumn{1}{c|}{\multirow{2}{*}{\textbf{$\dagger$MCS}}}    & \multirow{2}{*}{1.0659} & 1.0562              & 0.9882           & 0.8991                                                                & 1.0524          & 1.0611          & 1.0114              \\
\multicolumn{1}{c|}{}                                   & \multicolumn{1}{c|}{}                                 &                         & (-0.91\%)           & (-7.29\%)        & (-15.65\%)                                                            & (-1.27\%)       & (-0.45\%)       & (-5.11\%)           \\
\multicolumn{1}{c|}{\multirow{2}{*}{\textbf{English}}}  & \multicolumn{1}{c|}{\multirow{2}{*}{\textbf{*MCS}}}    & \multirow{2}{*}{1.0915} & 1.0919              & 1.0102           & 0.9839                                                                & 1.0952          & 1.0903          & 1.0543              \\
\multicolumn{1}{c|}{}                                   & \multicolumn{1}{c|}{}                                 &                         & (0.04\%)            & (-7.45\%)        & (-9.86\%)                                                             & (0.34\%)        & (-0.11\%)       & (-3.41\%)           \\
\multicolumn{1}{c|}{}                                   & \multicolumn{1}{c|}{\multirow{2}{*}{\textbf{PR-MCS}}} & \multirow{2}{*}{1.1484} & 0.7679              & 0.2721           & 0.0065                                                                & 0.0163          & 0.138           & 0.24016             \\
\multicolumn{1}{c|}{}                                   & \multicolumn{1}{c|}{}                                 &                         & (-33.13\%)          & (-76.31\%)       & (-99.43\%)                                                            & (-98.58\%)      & (-87.98\%)      & \textbf{(-79.09\%)} \\ \hline
\multicolumn{1}{c|}{\textbf{}}                          & \multicolumn{1}{c|}{\multirow{2}{*}{\textbf{$\dagger$MCS}}}    & \multirow{2}{*}{1.0688} & 1.067               & 0.9852           & 0.9059                                                                & 1.0738          & 1.0664          & 1.01966             \\
\multicolumn{1}{c|}{\textbf{}}                          & \multicolumn{1}{c|}{}                                 &                         & (-0.17\%)           & (-7.82\%)        & (-15.24\%)                                                            & (0.47\%)        & (-0.22\%)       & (-4.60\%)           \\
\multicolumn{1}{c|}{\multirow{2}{*}{\textbf{German}}}   & \multicolumn{1}{c|}{\multirow{2}{*}{\textbf{*MCS}}}    & \multirow{2}{*}{1.0779} & 1.0763              & 0.9958           & 0.9921                                                                & 1.0839          & 1.0796          & 1.04554             \\
\multicolumn{1}{c|}{}                                   & \multicolumn{1}{c|}{}                                 &                         & (-0.15\%)           & (-7.62\%)        & (-7.96\%)                                                             & (0.56\%)        & (0.16\%)        & (-3.00\%)           \\
\multicolumn{1}{c|}{\textbf{}}                          & \multicolumn{1}{c|}{\multirow{2}{*}{\textbf{PR-MCS}}} & \multirow{2}{*}{1.1621} & 0.7511              & 0.3756           & 0.0203                                                                & 0.0346          & 0.1556          & 0.26744             \\
\multicolumn{1}{c|}{\textbf{}}                          & \multicolumn{1}{c|}{}                                 &                         & (-35.37\%)          & (-67.68\%)       & (-98.25\%)                                                            & (-97.02\%)      & (-86.61\%)      & \textbf{(-76.99\%)} \\ \hline
\multicolumn{1}{c|}{\textbf{}}                          & \multicolumn{1}{c|}{\multirow{2}{*}{\textbf{$\dagger$MCS}}}    & \multirow{2}{*}{1.0653} & 1.0636              & 0.9815           & 0.8863                                                                & 1.0625          & 1.0594          & 1.01066             \\
\multicolumn{1}{c|}{\textbf{}}                          & \multicolumn{1}{c|}{}                                 &                         & (-0.16\%)           & (-7.87\%)        & (-16.80\%)                                                            & (-0.26\%)       & (-0.55\%)       & (-5.13\%)           \\
\multicolumn{1}{c|}{\multirow{2}{*}{\textbf{French}}}   & \multicolumn{1}{c|}{\multirow{2}{*}{\textbf{*MCS}}}    & \multirow{2}{*}{1.0758} & 1.0745              & 0.9986           & 0.9855                                                                & 1.0824          & 1.0778          & 1.04376             \\
\multicolumn{1}{c|}{}                                   & \multicolumn{1}{c|}{}                                 &                         & (-0.12\%)           & (-7.18\%)        & (-8.39\%)                                                             & (0.61\%)        & (0.19\%)        & (-2.98\%)           \\
\multicolumn{1}{c|}{\textbf{}}                          & \multicolumn{1}{c|}{\multirow{2}{*}{\textbf{PR-MCS}}} & \multirow{2}{*}{1.6681} & 1.0369              & 0.6367           & 0.018                                                                 & 0.0545          & 0.3915          & 0.42752             \\
\multicolumn{1}{c|}{\textbf{}}                          & \multicolumn{1}{c|}{}                                 &                         & (-37.84\%)          & (-61.83\%)       & (-98.92\%)                                                            & (-96.73\%)      & (-76.53\%)      & \textbf{(-74.37\%)} \\ \hline
\multicolumn{1}{c|}{\textbf{}}                          & \multicolumn{1}{c|}{\multirow{2}{*}{\textbf{$\dagger$MCS}}}    & \multirow{2}{*}{1.0527} & 1.0517              & 0.9707           & 0.8773                                                                & 1.0445          & 1.0451          & 0.99786             \\
\multicolumn{1}{c|}{\textbf{}}                          & \multicolumn{1}{c|}{}                                 &                         & (-0.09\%)           & (-7.79\%)        & (-16.66\%)                                                            & (-0.78\%)       & (-0.72\%)       & (-5.21\%)           \\
\multicolumn{1}{c|}{\multirow{2}{*}{\textbf{Spanish}}}  & \multicolumn{1}{c|}{\multirow{2}{*}{\textbf{*MCS}}}    & \multirow{2}{*}{1.0747} & 1.0732              & 0.999            & 0.9829                                                                & 1.0802          & 1.0756          & 1.04218             \\
\multicolumn{1}{c|}{}                                   & \multicolumn{1}{c|}{}                                 &                         & (-0.14\%)           & (-7.04\%)        & (-8.54\%)                                                             & (0.51\%)        & (0.08\%)        & (-3.03\%)           \\
\multicolumn{1}{c|}{\textbf{}}                          & \multicolumn{1}{c|}{\multirow{2}{*}{\textbf{PR-MCS}}} & \multirow{2}{*}{1.6534} & 1.0358              & 0.6189           & 0.0111                                                                & 0.0587          & 0.618           & 0.4685              \\
\multicolumn{1}{c|}{\textbf{}}                          & \multicolumn{1}{c|}{}                                 &                         & (-37.35\%)          & (-62.57\%)       & (-99.33\%)                                                            & (-96.45\%)      & (-62.62\%)      & \textbf{(-71.66\%)} \\ \hline
\multicolumn{1}{c|}{\textbf{}}                          & \multicolumn{1}{c|}{\multirow{2}{*}{\textbf{$\dagger$MCS}}}    & \multirow{2}{*}{1.0703} & 1.0722              & 0.9875           & 0.8573                                                                & 1.0796          & 1.067           & 1.01272             \\
\multicolumn{1}{c|}{\textbf{}}                          & \multicolumn{1}{c|}{}                                 &                         & (0.18\%)            & (-7.74\%)        & (-19.90\%)                                                            & (0.87\%)        & (-0.31\%)       & (-5.38\%)           \\
\multicolumn{1}{c|}{\multirow{2}{*}{\textbf{Japanese}}} & \multicolumn{1}{c|}{\multirow{2}{*}{\textbf{*MCS}}}    & \multirow{2}{*}{1.0293} & 1.0275              & 0.9655           & 0.8932                                                                & 1.0278          & 1.0255          & 0.9879              \\
\multicolumn{1}{c|}{}                                   & \multicolumn{1}{c|}{}                                 &                         & (-0.17\%)           & (-6.20\%)        & (-13.22\%)                                                            & (-0.15\%)       & (-0.37\%)       & (-4.02\%)           \\
\multicolumn{1}{c|}{\textbf{}}                          & \multicolumn{1}{c|}{\multirow{2}{*}{\textbf{PR-MCS}}} & \multirow{2}{*}{1.6279} & 0.7931              & 0.6235           & 0.0106                                                                & 0.0894          & 0.715           & 0.44632             \\
\multicolumn{1}{c|}{}                                   & \multicolumn{1}{c|}{}                                 &                         & (-51.28\%)          & (-61.70\%)       & (-99.35\%)                                                            & (-94.51\%)      & (-56.08\%)      & \textbf{(-72.58\%)} \\ \hline
                                                        &                                                       &                         &                     &                  &                                                                       & \multicolumn{3}{l}{$\dagger$ : \citet{multilingualCLIP2022} based}                       \\
                                                        &                                                       &                         &                     &                  &                                                                       & \multicolumn{3}{l}{*: Our Multilingual CLIP based}     
\end{tabular}
}
\caption{Flickr8k results table.}
\label{table5}
\end{table*}
\begin{table*}[p]
\renewcommand{\arraystretch}{1.3}
\centering
\resizebox{\textwidth}{!}{
\begin{tabular}{ccccccccc}
\hline
\multicolumn{1}{c|}{\textbf{Language}}                  & \multicolumn{1}{c|}{\textbf{Metric}}                  & \textbf{Original}       & \textbf{Repetition} & \textbf{Removal} & \textbf{\begin{tabular}[c]{@{}c@{}}Masking\end{tabular}} & \textbf{Jumble} & \textbf{Substitution} & \textbf{Average}    \\ \hline
\multicolumn{1}{c|}{}                                   & \multicolumn{1}{c|}{\multirow{2}{*}{\textbf{$\dagger$MCS}}}    & \multirow{2}{*}{0.7217} & 0.7134              & 0.6763           & 0.625                                                                 & 0.7007          & 0.7114          & 0.68536             \\
\multicolumn{1}{c|}{}                                   & \multicolumn{1}{c|}{}                                 &                         & (-1.15\%)           & (-6.29\%)        & (-13.4\%)                                                             & (-2.91\%)       & (-1.43\%)       & (-5.04\%)           \\
\multicolumn{1}{c|}{\multirow{2}{*}{\textbf{English}}}  & \multicolumn{1}{c|}{\multirow{2}{*}{\textbf{*MCS}}}    & \multirow{2}{*}{1.0626} & 1.0606              & 1.0063           & 0.9756                                                                & 1.0651          & 1.0614          & 1.0338              \\
\multicolumn{1}{c|}{}                                   & \multicolumn{1}{c|}{}                                 &                         & (-0.19\%)           & (-5.3\%)         & (-8.19\%)                                                             & (0.24\%)        & (-0.11\%)       & (-2.71\%)           \\
\multicolumn{1}{c|}{}                                   & \multicolumn{1}{c|}{\multirow{2}{*}{\textbf{PR-MCS}}} & \multirow{2}{*}{0.9769} & 0.6281              & 0.2754           & 0.0069                                                                & 0.0892          & 0.4726          & \textbf{0.29444}    \\
\multicolumn{1}{c|}{}                                   & \multicolumn{1}{c|}{}                                 &                         & (-35.7\%)           & (-71.81\%)       & (-99.29\%)                                                            & (-90.87\%)      & (-51.62\%)      & \textbf{(-69.86\%)} \\ \hline
\multicolumn{1}{c|}{\textbf{}}                          & \multicolumn{1}{c|}{\multirow{2}{*}{\textbf{$\dagger$MCS}}}    & \multirow{2}{*}{1.0284} & 1.0213              & 0.9757           & 0.9176                                                                & 1.0352          & 1.0254          & 0.99504             \\
\multicolumn{1}{c|}{\textbf{}}                          & \multicolumn{1}{c|}{}                                 &                         & (-0.69\%)           & (-5.12\%)        & (-10.77\%)                                                            & (0.66\%)        & (-0.29\%)       & (-3.24\%)           \\
\multicolumn{1}{c|}{\multirow{2}{*}{\textbf{German}}}   & \multicolumn{1}{c|}{\multirow{2}{*}{\textbf{*MCS}}}    & \multirow{2}{*}{1.0491} & 1.0433              & 0.9952           & 0.9848                                                                & 1.0515          & 0.9754          & 1.01004             \\
\multicolumn{1}{c|}{}                                   & \multicolumn{1}{c|}{}                                 &                         & (-0.55\%)           & (-5.14\%)        & (-6.13\%)                                                             & (0.23\%)        & (-7.03\%)       & (-3.72\%)           \\
\multicolumn{1}{c|}{\textbf{}}                          & \multicolumn{1}{c|}{\multirow{2}{*}{\textbf{PR-MCS}}} & \multirow{2}{*}{0.9895} & 0.6857              & 0.4573           & 0.0154                                                                & 0.1558          & 0.7607          & \textbf{0.41498}    \\
\multicolumn{1}{c|}{}                                   & \multicolumn{1}{c|}{}                                 &                         & (-30.7\%)           & (-53.78\%)       & (-98.44\%)                                                            & (-84.25\%)      & (-23.12\%)      & \textbf{(-58.06\%)} \\ \hline
\multicolumn{1}{c|}{\textbf{}}                          & \multicolumn{1}{c|}{\multirow{2}{*}{\textbf{$\dagger$MCS}}}    & \multirow{2}{*}{1.0266} & 1.0206              & 0.9788           & 0.9034                                                                & 1.0369          & 1.0228          & 0.9925              \\
\multicolumn{1}{c|}{\textbf{}}                          & \multicolumn{1}{c|}{}                                 &                         & (-0.58\%)           & (-4.66\%)        & (-12\%)                                                               & (1\%)           & (-0.37\%)       & (-3.32\%)           \\
\multicolumn{1}{c|}{\multirow{2}{*}{\textbf{French}}}   & \multicolumn{1}{c|}{\multirow{2}{*}{\textbf{*MCS}}}    & \multirow{2}{*}{1.0509} & 1.0456              & 1.0001           & 0.9764                                                                & 1.0534          & 1.0499          & 1.02508             \\
\multicolumn{1}{c|}{}                                   & \multicolumn{1}{c|}{}                                 &                         & (-0.5\%)            & (-4.83\%)        & (-7.09\%)                                                             & (0.24\%)        & (-0.1\%)        & (-2.46\%)           \\
\multicolumn{1}{c|}{\textbf{}}                          & \multicolumn{1}{c|}{\multirow{2}{*}{\textbf{PR-MCS}}} & \multirow{2}{*}{1.442}  & 0.9655              & 0.7088           & 0.0097                                                                & 0.2314          & 1.166           & \textbf{0.61628}    \\
\multicolumn{1}{c|}{}                                   & \multicolumn{1}{c|}{}                                 &                         & (-33.04\%)          & (-50.85\%)       & (-99.33\%)                                                            & (-83.95\%)      & (-19.14\%)      & \textbf{(-57.26\%)} \\ \hline
\multicolumn{1}{c|}{}                                   & \multicolumn{1}{c|}{\multirow{2}{*}{\textbf{$\dagger$MCS}}}    & \multirow{2}{*}{1.0226} & 1.0189              & 0.9794           & 0.8924                                                                & 1.0377          & 1.0207          & 0.98982             \\
\multicolumn{1}{c|}{\textbf{}}                          & \multicolumn{1}{c|}{}                                 &                         & (-0.36\%)           & (-4.22\%)        & (-12.73\%)                                                            & (1.48\%)        & (-0.19\%)       & (-3.21\%)           \\
\multicolumn{1}{c|}{\multirow{2}{*}{\textbf{Spanish}}}  & \multicolumn{1}{c|}{\multirow{2}{*}{\textbf{*MCS}}}    & \multirow{2}{*}{1.0505} & 1.0463              & 1.002            & 0.973                                                                 & 1.0537          & 0.9982          & 1.01464             \\
\multicolumn{1}{c|}{}                                   & \multicolumn{1}{c|}{}                                 &                         & (-0.4\%)            & (-4.62\%)        & (-7.38\%)                                                             & (0.3\%)         & (-4.98\%)       & (-3.41\%)           \\
\multicolumn{1}{c|}{\textbf{}}                          & \multicolumn{1}{c|}{\multirow{2}{*}{\textbf{PR-MCS}}} & \multirow{2}{*}{1.4526} & 0.9893              & 0.6767           & 0.003                                                                 & 0.209           & 1.2505          & \textbf{0.6257}     \\
\multicolumn{1}{c|}{}                                   & \multicolumn{1}{c|}{}                                 &                         & (-31.89\%)          & (-53.41\%)       & (-99.79\%)                                                            & (-85.61\%)      & (-13.91\%)      & \textbf{(-56.93\%)} \\ \hline
\multicolumn{1}{c|}{\textbf{}}                          & \multicolumn{1}{c|}{\multirow{2}{*}{\textbf{$\dagger$MCS}}}    & \multirow{2}{*}{1.0297} & 1.0279              & 0.9855           & 0.8733                                                                & 1.05            & 1.0259          & 0.99252             \\
\multicolumn{1}{c|}{\textbf{}}                          & \multicolumn{1}{c|}{}                                 &                         & (-0.17\%)           & (-4.29\%)        & (-15.19\%)                                                            & (1.97\%)        & (-0.37\%)       & (-3.61\%)           \\
\multicolumn{1}{c|}{\multirow{2}{*}{\textbf{Japanese}}} & \multicolumn{1}{c|}{\multirow{2}{*}{\textbf{*MCS}}}    & \multirow{2}{*}{1.0414} & 1.0242              & 0.9963           & 0.9275                                                                & 1.0406          & 1.0403          & 1.00578             \\
\multicolumn{1}{c|}{}                                   & \multicolumn{1}{c|}{}                                 &                         & (-1.65\%)           & (-4.33\%)        & (-10.94\%)                                                            & (-0.08\%)       & (-0.11\%)       & (-3.42\%)           \\
\multicolumn{1}{c|}{\textbf{}}                          & \multicolumn{1}{c|}{\multirow{2}{*}{\textbf{PR-MCS}}} & \multirow{2}{*}{1.4113} & 0.8612              & 0.7769           & 0.0035                                                                & 0.3241          & 1.1465          & \textbf{0.62244}    \\
\multicolumn{1}{c|}{}                                   & \multicolumn{1}{c|}{}                                 &                         & (-38.98\%)          & (-44.95\%)       & (-99.75\%)                                                            & (-77.04\%)      & (-18.76\%)      & \textbf{(-55.9\%)}  \\ \hline
                                                        &                                                       &                         &                     &                  &                                                                       & \multicolumn{3}{l}{$\dagger$ : \citet{multilingualCLIP2022} based}                       \\
                                                        &                                                       &                         &                     &                  &                                                                       & \multicolumn{3}{l}{*: Our Multilingual CLIP based}     
\end{tabular}
}
\caption{VizWiz results table.}
\label{table6}
\end{table*}
\begin{table*}[p]
\renewcommand{\arraystretch}{1.3}
\centering
\resizebox{\textwidth}{!}{
\begin{tabular}{ccccccccc}
\hline
\multicolumn{1}{c|}{\textbf{Language}}                  & \multicolumn{1}{c|}{\textbf{Metric}}                                                                       & \textbf{Original}       & \textbf{Repetition} & \textbf{Removal} & \textbf{\begin{tabular}[c]{@{}c@{}}Masking\end{tabular}} & \textbf{Jumble} & \textbf{Substitution} & \textbf{Average}    \\ \hline
\multicolumn{1}{c|}{\textbf{}}                          & \multicolumn{1}{c|}{\multirow{2}{*}{\textbf{$\dagger$MCS}}}                                                         & \multirow{2}{*}{0.7593} & 0.7524              & 0.6577           & 0.6265                                                                & 0.7275          & 0.7515          & 0.70312             \\
\multicolumn{1}{c|}{}                                   & \multicolumn{1}{c|}{}                                                                                      &                         & (-0.91\%)           & (-13.38\%)       & (-17.49\%)                                                            & (-4.19\%)       & (-1.03\%)       & (-7.40\%)           \\
\multicolumn{1}{c|}{}                                   & \multicolumn{1}{c|}{\multirow{2}{*}{\textbf{*MCS}}}                                                        & \multirow{2}{*}{1.0657} & 1.0439              & 0.9699           & 1.0484                                                                & 1.0628          & 1.061           & 1.0372              \\
\multicolumn{1}{c|}{\multirow{2}{*}{\textbf{English}}}  & \multicolumn{1}{c|}{}                                                                                      &                         & (-2.05\%)           & (-8.99\%)        & (-1.62\%)                                                             & (-0.27\%)       & (-0.44\%)       & (-2.67\%)           \\
\multicolumn{1}{c|}{}                                   & \multicolumn{1}{c|}{\multirow{2}{*}{\textbf{PR-MCS}}}                                                      & \multirow{2}{*}{1.0429} & 0.5984              & 0.2382           & 0.4844                                                                & 0.7262          & 0.7421          & \textbf{0.55786}    \\
\multicolumn{1}{c|}{}                                   & \multicolumn{1}{c|}{}                                                                                      &                         & (-42.62\%)          & (-77.16\%)       & (-53.55\%)                                                            & (-30.37\%)      & (-28.84\%)      & \textbf{(-46.51\%)} \\
\multicolumn{1}{c|}{}                                   & \multicolumn{1}{c|}{\multirow{2}{*}{\textbf{\begin{tabular}[c]{@{}c@{}}PR-MCS\\ (Few-shot)\end{tabular}}}} & \multirow{2}{*}{0.7125} & 0.3183              & 0.0653           & 0.0012                                                                & 0.0027          & 0.4045          & \textbf{0.1584}     \\
\multicolumn{1}{c|}{}                                   & \multicolumn{1}{c|}{}                                                                                      &                         & (-55.33\%)          & (-90.84\%)       & (-99.83\%)                                                            & (-99.62\%)      & (-43.23\%)      & \textbf{(-77.77\%)} \\ \hline
\multicolumn{1}{c|}{\textbf{}}                          & \multicolumn{1}{c|}{\multirow{2}{*}{\textbf{$\dagger$MCS}}}                                                         & \multirow{2}{*}{1.0708} & 1.0576              & 0.9516           & 1.0011                                                                & 1.0546          & 1.0643          & 1.02584             \\
\multicolumn{1}{c|}{}                                   & \multicolumn{1}{c|}{}                                                                                      &                         & (-1.23\%)           & (-11.13\%)       & (-6.51\%)                                                             & (-1.51\%)       & (-0.61\%)       & (-4.20\%)           \\
\multicolumn{1}{c|}{\textbf{}}                          & \multicolumn{1}{c|}{\multirow{2}{*}{\textbf{*MCS}}}                                                        & \multirow{2}{*}{1.0803} & 1.063               & 0.9638           & 1.0484                                                                & 1.0815          & 1.0781          & 1.04696             \\
\multicolumn{1}{c|}{\multirow{2}{*}{\textbf{German}}}   & \multicolumn{1}{c|}{}                                                                                      &                         & (-1.60\%)           & (-10.78\%)       & (-2.95\%)                                                             & (0.11\%)        & (-0.20\%)       & (-3.09\%)           \\
\multicolumn{1}{c|}{}                                   & \multicolumn{1}{c|}{\multirow{2}{*}{\textbf{PR-MCS}}}                                                      & \multirow{2}{*}{0.7261} & 0.5437              & 0.2521           & 0.4996                                                                & 0.5151          & 0.495           & \textbf{0.4611}     \\
\multicolumn{1}{c|}{}                                   & \multicolumn{1}{c|}{}                                                                                      &                         & (-25.12\%)          & (-65.28\%)       & (-31.19\%)                                                            & (-29.06\%)      & (-31.83\%)      & \textbf{(-36.5\%)}  \\
\multicolumn{1}{c|}{}                                   & \multicolumn{1}{c|}{\multirow{2}{*}{\textbf{\begin{tabular}[c]{@{}c@{}}PR-MCS\\ (Few-shot)\end{tabular}}}} & \multirow{2}{*}{0.589}  & 0.396               & 0.0617           & 0.0011                                                                & 0.1059          & 0.3062          & \textbf{0.17418}    \\
\multicolumn{1}{c|}{}                                   & \multicolumn{1}{c|}{}                                                                                      &                         & (-32.77\%)          & (-89.52\%)       & (-99.81\%)                                                            & (-82.02\%)      & (-48.01\%)      & \textbf{(-70.43\%)} \\ \hline
\multicolumn{1}{c|}{\textbf{}}                          & \multicolumn{1}{c|}{\multirow{2}{*}{\textbf{$\dagger$MCS}}}                                                         & \multirow{2}{*}{1.053}  & 1.045               & 0.9462           & 0.991                                                                 & 1.048           & 1.0463          & 1.0153              \\
\multicolumn{1}{c|}{}                                   & \multicolumn{1}{c|}{}                                                                                      &                         & (-0.76\%)           & (-10.14\%)       & (-5.89\%)                                                             & (-0.47\%)       & (-0.64\%)       & (-3.58\%)           \\
\multicolumn{1}{c|}{\textbf{}}                          & \multicolumn{1}{c|}{\multirow{2}{*}{\textbf{*MCS}}}                                                        & \multirow{2}{*}{1.0575} & 1.0472              & 0.957            & 1.0322                                                                & 1.0574          & 1.0536          & 1.02948             \\
\multicolumn{1}{c|}{\multirow{2}{*}{\textbf{French}}}   & \multicolumn{1}{c|}{}                                                                                      &                         & (-0.97\%)           & (-9.50\%)        & (-2.39\%)                                                             & (-0.01\%)       & (-0.37\%)       & (-2.65\%)           \\
\multicolumn{1}{c|}{}                                   & \multicolumn{1}{c|}{\multirow{2}{*}{\textbf{PR-MCS}}}                                                      & \multirow{2}{*}{1.3634} & 0.96                & 0.4739           & 1.0976                                                                & 0.7335          & 0.9791          & \textbf{0.84882}    \\
\multicolumn{1}{c|}{}                                   & \multicolumn{1}{c|}{}                                                                                      &                         & (-29.59\%)          & (-65.24\%)       & (-19.50\%)                                                            & (-46.2\%)       & (-28.19\%)      & \textbf{(-37.74\%)} \\
\multicolumn{1}{c|}{}                                   & \multicolumn{1}{c|}{\multirow{2}{*}{\textbf{\begin{tabular}[c]{@{}c@{}}PR-MCS\\ (Few-shot)\end{tabular}}}} & \multirow{2}{*}{1.4862} & 1.025               & 0.1559           & 0.0082                                                                & 0.1034          & 0.9032          & \textbf{0.43914}    \\
\multicolumn{1}{c|}{}                                   & \multicolumn{1}{c|}{}                                                                                      &                         & (-31.03\%)          & (-89.51\%)       & (-99.45\%)                                                            & (-93.04\%)      & (-39.23\%)      & \textbf{(-70.45\%)} \\ \hline
\multicolumn{1}{c|}{\textbf{}}                          & \multicolumn{1}{c|}{\multirow{2}{*}{\textbf{$\dagger$MCS}}}                                                         & \multirow{2}{*}{1.0599} & 1.048               & 0.9548           & 0.9996                                                                & 1.0464          & 1.0536          & 1.02048             \\
\multicolumn{1}{c|}{}                                   & \multicolumn{1}{c|}{}                                                                                      &                         & (-1.12\%)           & (-9.92\%)        & (-5.69\%)                                                             & (-1.27\%)       & (-0.59\%)       & (-3.72\%)           \\
\multicolumn{1}{c|}{\textbf{}}                          & \multicolumn{1}{c|}{\multirow{2}{*}{\textbf{*MCS}}}                                                        & \multirow{2}{*}{1.0577} & 1.0385              & 0.9629           & 1.0393                                                                & 1.0542          & 1.0537          & 1.02972             \\
\multicolumn{1}{c|}{\multirow{2}{*}{\textbf{Spanish}}}  & \multicolumn{1}{c|}{}                                                                                      &                         & (-1.82\%)           & (-8.96\%)        & (-1.74\%)                                                             & (-0.33\%)       & (-0.38\%)       & (-2.65\%)           \\
\multicolumn{1}{c|}{}                                   & \multicolumn{1}{c|}{\multirow{2}{*}{\textbf{PR-MCS}}}                                                      & \multirow{2}{*}{1.3821} & 1.206               & 0.5811           & 1.1648                                                                & 0.9128          & 0.8481          & \textbf{0.94256}    \\
\multicolumn{1}{c|}{}                                   & \multicolumn{1}{c|}{}                                                                                      &                         & (-12.74\%)          & (-57.96\%)       & (-15.72\%)                                                            & (-33.96\%)      & (-38.64\%)      & \textbf{(-31.80\%)} \\
\multicolumn{1}{c|}{}                                   & \multicolumn{1}{c|}{\multirow{2}{*}{\textbf{\begin{tabular}[c]{@{}c@{}}PR-MCS\\ (Few-shot)\end{tabular}}}} & \multirow{2}{*}{1.4292} & 1.187               & 0.1798           & 0.0019                                                                & 0.1405          & 0.473           & \textbf{0.39644}    \\
\multicolumn{1}{c|}{}                                   & \multicolumn{1}{c|}{}                                                                                      &                         & (-16.95\%)          & (-87.42\%)       & (-99.87\%)                                                            & (-90.17\%)      & (-66.90\%)      & \textbf{(-72.26\%)} \\ \hline
\multicolumn{1}{c|}{\textbf{}}                          & \multicolumn{1}{c|}{\multirow{2}{*}{\textbf{$\dagger$MCS}}}                                                         & \multirow{2}{*}{1.06}   & 1.0546              & 0.9768           & 0.9976                                                                & 1.0475          & 1.0559          & 1.02648             \\
\multicolumn{1}{c|}{}                                   & \multicolumn{1}{c|}{}                                                                                      &                         & (-0.51\%)           & (-7.85\%)        & (-5.89\%)                                                             & (-1.18\%)       & (-0.39\%)       & (-3.16\%)           \\
\multicolumn{1}{c|}{\textbf{}}                          & \multicolumn{1}{c|}{\multirow{2}{*}{\textbf{*MCS}}}                                                        & \multirow{2}{*}{0.963}  & 0.9692              & 0.982            & 1.0323                                                                & 0.9813          & 0.963           & 0.98556             \\
\multicolumn{1}{c|}{\multirow{2}{*}{\textbf{Japanese}}} & \multicolumn{1}{c|}{}                                                                                      &                         & (0.64\%)            & (1.97\%)         & (7.20\%)                                                              & (1.90\%)        & (0.00\%)        & (2.34\%)            \\
\multicolumn{1}{c|}{}                                   & \multicolumn{1}{c|}{\multirow{2}{*}{\textbf{PR-MCS}}}                                                      & \multirow{2}{*}{1.1551} & 0.2719              & 0.3357           & 0.7354                                                                & 0.6373          & 1.0862          & \textbf{0.6133}     \\
\multicolumn{1}{c|}{}                                   & \multicolumn{1}{c|}{}                                                                                      &                         & (-76.46\%)          & (-70.94\%)       & (-36.33\%)                                                            & (-44.83\%)      & (-5.96\%)       & \textbf{(-46.91\%)} \\
\multicolumn{1}{c|}{}                                   & \multicolumn{1}{c|}{\multirow{2}{*}{\textbf{\begin{tabular}[c]{@{}c@{}}PR-MCS\\ (Few-shot)\end{tabular}}}} & \multirow{2}{*}{1.4136} & 0.1385              & 0.1285           & 0.0012                                                                & 0.0048          & 1.2969          & \textbf{0.31398}    \\
\multicolumn{1}{c|}{\textbf{}}                          & \multicolumn{1}{c|}{}                                                                                      &                         & (-90.2\%)           & (-90.91\%)       & (-99.92\%)                                                            & (-99.66\%)      & (-8.26\%)       & \textbf{(-77.79\%)} \\ \hline
\textbf{}                                               & \textbf{}                                                                                                  &                         &                     &                  &                                                                       & \multicolumn{3}{l}{$\dagger$ : \citet{multilingualCLIP2022} based}                       \\
\textbf{}                                               & \textbf{}                                                                                                  &                         &                     &                  &                                                                       & \multicolumn{3}{l}{*: Our Multilingual CLIP based}     
\end{tabular}
}
\caption{M-FineCapEval results table.}
\label{table7}
\end{table*}

\end{document}